\documentclass{article}

\PassOptionsToPackage{numbers, compress}{natbib}
\usepackage[preprint]{neurips_2026}

\usepackage[utf8]{inputenc} 
\usepackage[T1]{fontenc}    
\usepackage{hyperref}       
\usepackage{url}            
\usepackage{booktabs}       
\usepackage{amsfonts}       
\usepackage{nicefrac}       
\usepackage{microtype}      
\usepackage{xcolor}         
\usepackage{amsmath}
\usepackage{graphicx}
\usepackage{multirow}

\title{EquiFiLM: Charge-Conditioned Equivariant Force Fields via Feature-wise Linear Modulation}

\author{%
Samuel Sahel-Schackis$^{1,2,3*}$,\quad Ken-ichi Nomura$^{4}$,\quad Aiichiro Nakano$^{4}$,\\
\textbf{Matthias F. Kling}$^{3,5}$,\quad \textbf{Thomas Linker}$^{2,3}$\\
\\
$^1$Department of Physics, Stanford University \\
$^2$Linac Coherent Light Source, SLAC National Accelerator Laboratory \\
$^3$Stanford PULSE Institute, SLAC National Accelerator Laboratory \\ 
$^4$Collaboratory for Advanced Computing and Simulation, University of Southern California \\
$^5$Department of Applied Physics, Stanford University \\
\\
$^*$\texttt{samss@slac.stanford.edu}
}

\begin{document}

\maketitle

\begin{abstract}
  Foundation machine learning force fields (MLFFs) such as MACE-MP-0 and UMA cover broad chemical space at near density functional theory (DFT) accuracy. However, they assume equilibrium ground-state physics and do not natively handle externally induced changes to the electronic state, such as charging, applied fields, or electronic excitation, which limits their use for driven processes such as photoexcitation and charge injection. We propose EquiFiLM, a lightweight extension that adds continuous external conditioning to any equivariant foundation MLFF via a per-layer Feature-wise Linear Modulation (FiLM) block, learning externally driven changes to the potential energy surface from minimal training data. The block modulates only scalar channels and preserves E(3)-equivariance exactly. We demonstrate the recipe on charged liquid water with the foundation model MACE-MatPES as the backbone, yielding E-MACE. On the four training charges, E-MACE delivers a $3.1\times$ reduction in force RMSE ($21.3$ to $6.96$ meV/\AA{}) and a $61\times$ reduction in per-atom energy RMSE ($6.1$ to $0.1$ meV/atom) over a baseline without EquiFiLM trained on the same data, at indistinguishable inference cost. Across seven held-out interpolation and extrapolation charges, force RMSE stays within $18-61$ meV/\AA{} and energy RMSE within $0.7-5.4$ meV/atom. The model runs stable molecular dynamics across the full range tested and predicts the charge-dependent first-shell response of the reduced pair distribution function probed by ultrafast electron diffraction. Adding this conditioning axis to the foundation requires only a few thousand DFT-labeled frames, against the $\approx 10^8$ structures of a charge-aware foundation trained from scratch. The recipe is backbone- and conditioning-agnostic: it applies without architectural change to any equivariant MLFF with scalar interaction-layer channels.
\end{abstract}

\clearpage

\section{Introduction}

Atomistic simulation under continuous external control, conditioned by total charge, temperature, applied field, hydrostatic pressure or doping fraction, is central to electrochemistry, photochemistry, doped-materials design and many other practical applications. Foundation machine-learning force fields (MLFFs) such as MACE-MP-0 \citep{batatia2024macemp} and UMA \citep{wood2025uma} now reach near-DFT accuracy across most of the periodic table, alongside other state-of-the-art interatomic potentials including the strictly equivariant architectures MACE-MatPES \citep{kaplan2025MatPES}, EquiformerV3 \citep{liao2026equiformerv3}, eSEN \citep{fu2025esen} and SevenNet-Omni \citep{kim2025sevennetomni}, and the non-equivariant or rotationally-unconstrained alternatives Orb-v3 \citep{rhodes2025orbv3} and PET-MAD \citep{mazitov2025petmad}. In their standard formulations, these condition exclusively on atomic positions and species. The model has no input pathway through which a per-graph external scalar can propagate, so two systems that share a geometry but differ in their conditioning produce bit-identical predictions.

\begin{figure}[ht]
  \centering
  \includegraphics[width=0.8\linewidth]{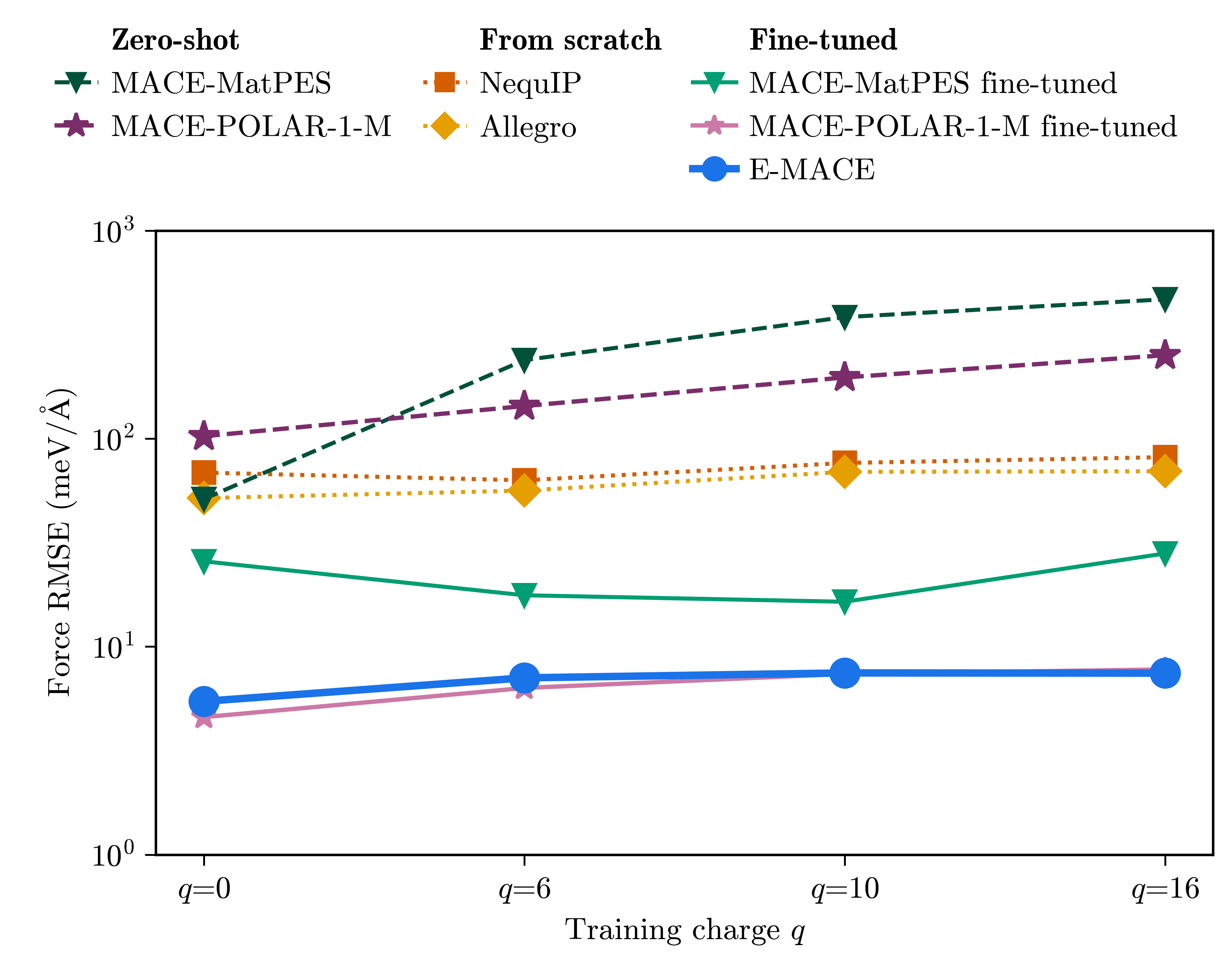}
  \caption{Per-charge force RMSE at the four training charges for seven models across three regimes: \emph{zero-shot foundations} (dashed), \emph{trained from scratch} (dotted), and \emph{fine-tuned on the training data} (solid). E-MACE improves greatly on its MACE-MatPES backbone, and reaches the accuracy of the far larger, purpose-built charge-aware foundation MACE-POLAR-1-M fine-tuned.}
  \label{fig:model_comparison}
\end{figure}

Current approaches to bringing external conditioning into MLFFs trade off generalization against added architectural and training cost. A foundation MLFF that ignores charge degrades quickly as charge is added, with force errors rising into the hundreds of meV/\AA{} on charged liquid water (Figure~\ref{fig:model_comparison}). Per-state specialists (one model per charge) reach acceptable accuracy on their training charge but cannot generalize to unseen charges (Appendix~\ref{app:crossmatrix}). Physics-grounded charge-aware foundations such as MACE-POLAR-1-M \citep{batatia2026macepolar1} and the broader charge-equilibration lineage of 4G-HDNNP \citep{ko20214ghdnnp}, AIMNet2-NSE \citep{kalita2025aimnet2nse}, DPLR \citep{zhang2022dplr} and SpookyNet \citep{unke2021spookynet} solve the charge case by construction. However, they do so at the steep cost of dedicated atomic-charge supervision, explicit per-atom Coulomb machinery or iterative charge equilibration, with foundation-scale members of this family trained from scratch on corpora of order $10^8$ structures \citep{batatia2026macepolar1}.

We propose \textbf{EquiFiLM}, a lightweight per-layer adapter that grants any equivariant foundation MLFF a continuous external-conditioning axis. The mechanism is a small Feature-wise Linear Modulation block \citep{perez2018film} whose gating parameters are produced by a per-graph multilayer perceptron (MLP) from the conditioning input. The block modulates only the scalar channels of every interaction layer, leaves the higher-order equivariant features untouched, and preserves E(3)-equivariance exactly.

Applied to the MACE-MatPES backbone on charged liquid water as the demonstration system, the resulting model E-MACE matches the best per-charge force root mean square error (RMSE) on the training charges of any method we compared, that of the far larger charge-aware foundation MACE-POLAR-1-M fine-tuned, and far surpasses the unconditioned trained foundations (Figure~\ref{fig:model_comparison}) at inference cost statistically indistinguishable from the same backbone fine-tuned without EquiFiLM (Table~\ref{tab:inference_time}). It generalizes to interpolation and extrapolation charges that it never saw in training on both force and energy, and runs stable molecular dynamics across the full charge range tested. Because the adapter is backbone-agnostic and modulates only scalar channels, the same construction extends in principle to other equivariant MLFFs and to other continuous scalar external variables such as temperature, hydrostatic pressure or doping fraction. 

\begin{table}[ht]
    \centering
    \caption{Single-step inference cost on a single A100 GPU for a $324$-atom cell at batch size $1$. The FiLM adapter does not add a measurable per-step cost relative to the unconditioned MACE-MatPES backbone fine-tuned on the same data, and is $3 \times$ faster than MACE-POLAR-1-M.}
    \begin{tabular}{lc}
         \toprule
         \textbf{Model} & \textbf{Inference time ($\mu$s / atom-step)} \\
         \midrule
         MACE-MatPES fine-tuned                & $191 \pm 1$ \\
         \textbf{E-MACE}                       & $\mathbf{193 \pm 2}$ \\
         MACE-MatPES                           & $213 \pm 3$ \\
         MACE-POLAR-1-M fine-tuned              & $578 \pm 5$ \\
         MACE-POLAR-1-M                         & $582 \pm 4$ \\
         \bottomrule
    \end{tabular}
    \label{tab:inference_time}
\end{table}

\section{Method}

\subsection{Equivariant message passing in MACE}

EquiFiLM is instantiated on top of an equivariant message-passing backbone, specifically MACE-MatPES \citep{kaplan2025MatPES}. The design choices below generalize to any equivariant interatomic potential with scalar channels at its interaction layers. Each atom $i$ in such a network carries a node feature $h_i^{(t)} = \bigoplus_{\ell\geq 0} h_i^{(t,\ell)}$ that decomposes into irreducible representations of $O(3)$ of order $\ell$: scalars ($\ell=0$), vectors ($\ell=1$), and higher-order tensors. Messages between atoms are constructed from tensor products of irreducible features along edges. MACE \citep{batatia2022mace} uses higher-body-order messages from the Atomic Cluster Expansion \citep{drautz2019ace} and reaches reference accuracy with only two interaction layers. Each interaction layer outputs a message tensor $m_i^{(t)} = \bigoplus_\ell m_i^{(t,\ell)}$ followed by a linear projection $h_i^{(t+1)} = W \cdot m_i^{(t)}$.

A function $f$ from positions and features to outputs is E(3)-equivariant if for any rotation $R\in O(3)$ and translation $t\in\mathbb{R}^3$,
\begin{equation}
f(R x + t,\; \rho_{\rm in}(R)\, h^{\rm in}) \;=\; \rho_{\rm out}(R)\, f(x,\, h^{\rm in})\,,
\end{equation}
where $\rho_{\ell}(R)$ acts on $h^{(\ell)}$ via the Wigner-$D$ matrix of order $\ell$. Equivariance is preserved under any operation that maps scalar features to scalar features and acts as the identity on $\ell\geq 1$ channels. Multiplying or shifting the scalar slice of a feature tensor by a function of a per-graph scalar therefore preserves E(3)-equivariance exactly, since that function is itself $O(3)$-invariant.

\subsection{ChargeFiLMBlock}

\begin{figure}[ht]
\centering
\includegraphics[width=0.8\linewidth]{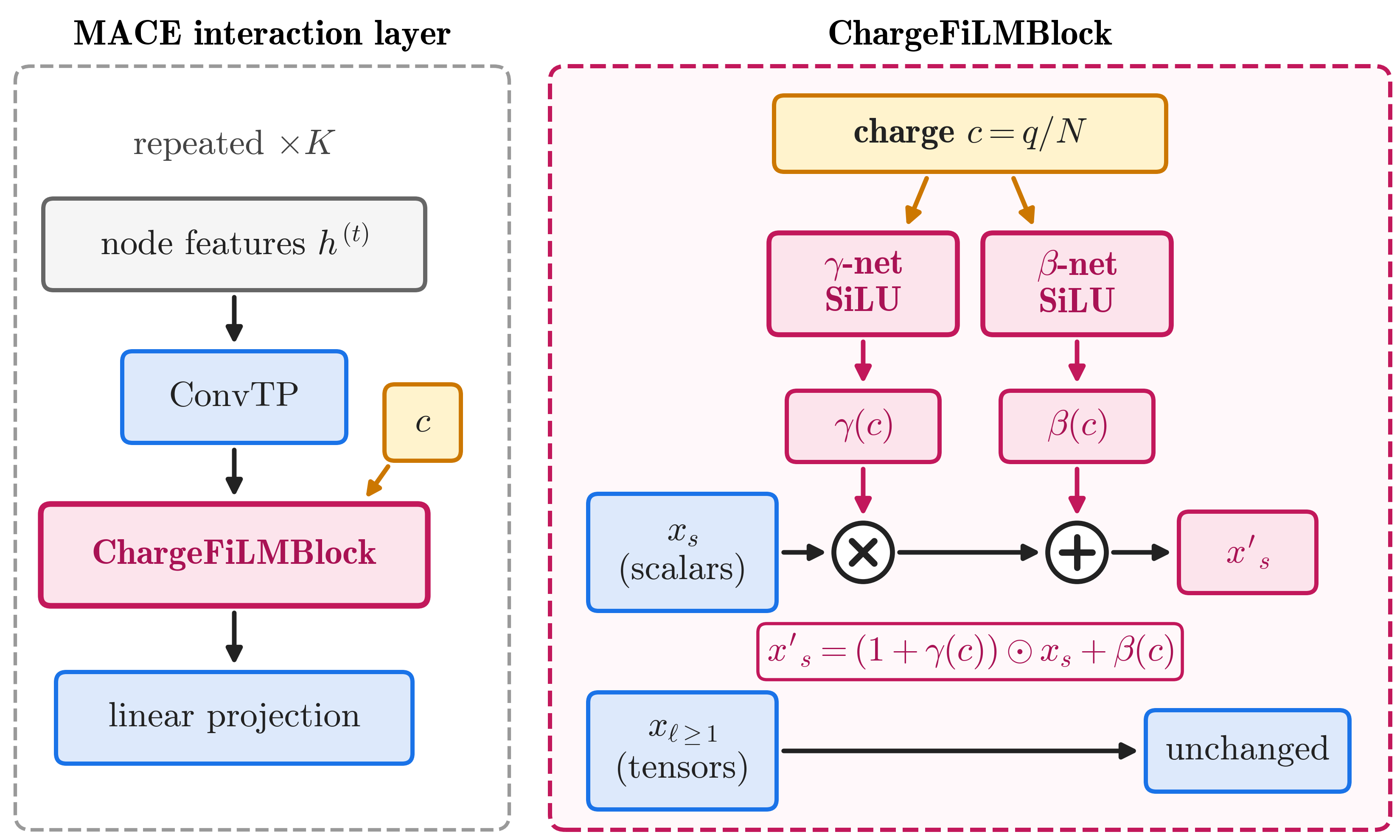}
\caption{ChargeFiLMBlock and its integration with MACE. \emph{Left}: a ChargeFiLMBlock inserted between the equivariant tensor product and the linear projection of a MACE interaction layer, applied independently at each of the $K=2$ layers. \emph{Right}: two small MLPs map the per-graph scalar $c$ to $\gamma(c)$ and $\beta(c)$, which scale and shift the scalar channels of the message tensor by $\mathbf{1}+\gamma(c)$ and $\beta(c)$ while the equivariant ($\ell\geq1$) channels pass through unchanged. The scalar channel count $N_s$ is $128$ and $256$ respectively for the two layers.}
\label{fig:architecture}
\end{figure}

Let $c\in\mathbb{R}$ be a per-graph conditioning scalar (in the experiments here the total charge $q$, entered as the per-atom density $c = q/N$ with $N$ the cell atom count) and let $N_s$ be the number of scalar channels in interaction layer $t$. Two independent two-layer MLPs are defined, each $\mathbb{R} \to \mathbb{R}^{N_s}$ with hidden width $h$ and SiLU activation, producing $\gamma(c) \in \mathbb{R}^{N_s}$ and $\beta(c) \in \mathbb{R}^{N_s}$ separately. The ChargeFiLMBlock transforms the scalar slice of every node feature in layer $t$ as
\begin{equation}
\label{eq:filmblock}
m_i^{\prime\,(t,\ell=0)} \;=\; \big(\mathbf{1} + \gamma(c)\big) \,\odot\, m_i^{(t,\ell=0)} \;+\; \beta(c)\,,
\qquad
m_i^{\prime\,(t,\ell\geq 1)} \;=\; m_i^{(t,\ell\geq 1)}\,,
\end{equation}
where $\odot$ denotes element-wise multiplication and $\mathbf{1}$ the vector of ones. The block is inserted between the message tensor $m^{(t)}$ and the linear projection $W$, as illustrated in Figure~\ref{fig:architecture}. The gating parameters $\gamma$ and $\beta$ depend only on the per-graph scalar $c$ and act only on scalar channels, so the block preserves E(3)-equivariance. Atom-permutation equivariance follows because $c$ is identical for every node.

The output linear layers of the two gating MLPs are initialized to zero so that $\gamma(c)\equiv\beta(c)\equiv \mathbf{0}$ at the start of training and the block reduces to the identity, so the adapter is bit-identical to the backbone it is attached to at the start of adapter training. The adapter adds approximately $15\%$ parameters relative to the MACE-MatPES backbone ($\approx 0.10$ M adapter vs. $\approx 0.66$ M backbone) and incurs no measurable per-step inference cost (Table~\ref{tab:inference_time}, more details in Appendix~\ref{app:filmcost}).

\subsection{Integration and training}

Training labels come from ab initio molecular dynamics (AIMD) simulations of liquid water at four total electronic charges $q\in\{0,6e,10e,16e\}$, where $q$ denotes the number of electrons added to a closed-shell neutral reference cell, with energies and forces computed by VASP \citep{kresse1996vasp} at the r$^2$SCAN meta-GGA level \citep{furness2020r2scan} (full DFT settings in Appendix~\ref{app:dataset}). The four trajectories supply $\approx 6,400$ frames. A uniform $10\%$ random split within each trajectory is held out as a validation set and never enters any gradient update. All training charge accuracy numbers reported below are evaluated on this split at the final stochastic-weight-averaging \citep{izmailov2018swa} epoch (further parameters in Appendix~\ref{app:hparams}).

\section{Results}

\subsection{Charge generalization}

\begin{figure}[ht]
\centering
\includegraphics[width=0.8\linewidth]{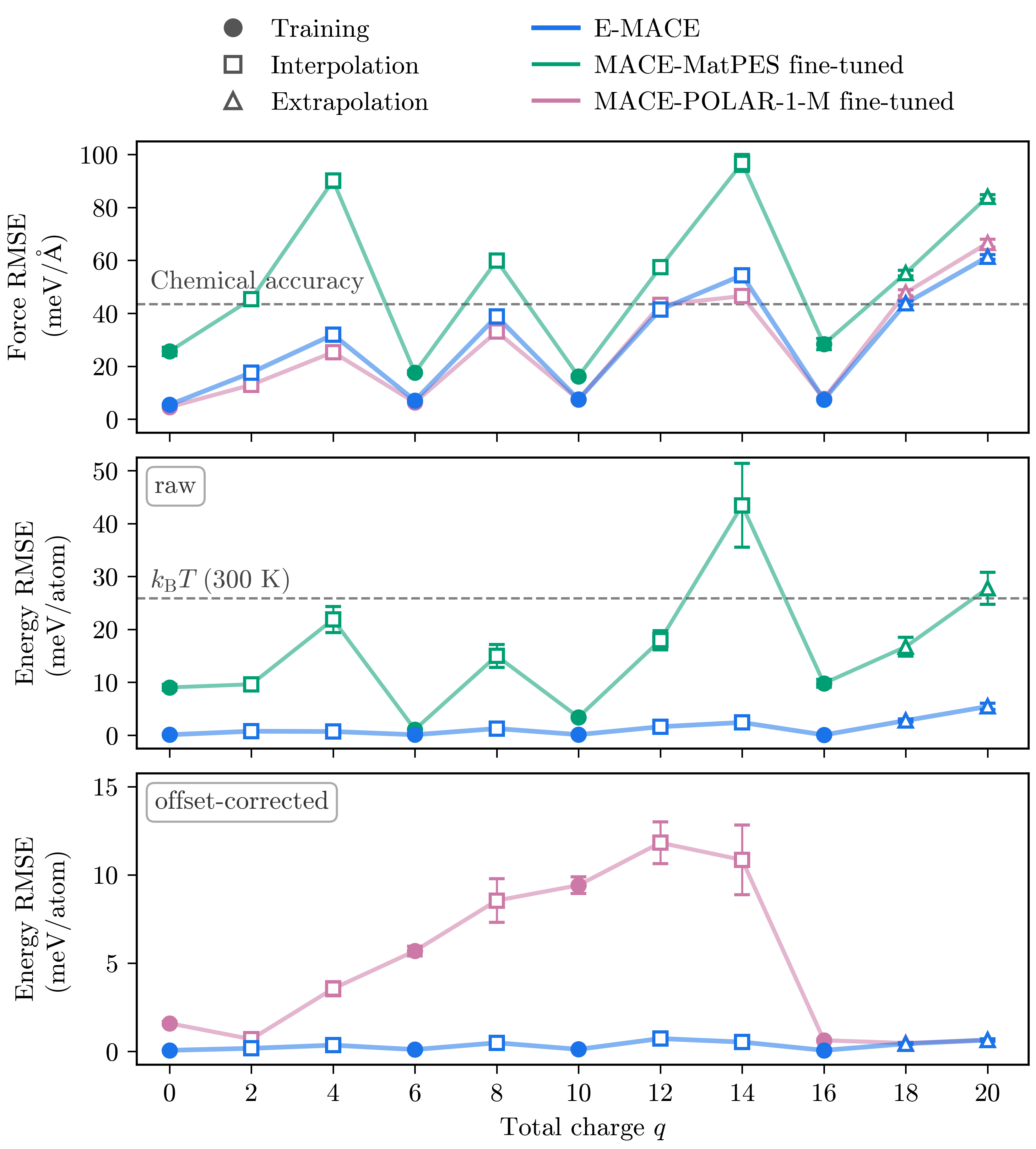}
\caption{Per-charge RMSE against VASP r$^2$SCAN labels; marker shape gives the split (circle: training, square: interpolation, triangle: extrapolation). \emph{Top:} force RMSE for E-MACE, the no-FiLM baseline (or equivalently fine-tuned MACE-MatPES), and fine-tuned MACE-POLAR-1-M (dashed: chemical accuracy, $1$ kcal/mol/\AA{}). \emph{Middle:} raw per-atom energy vs. the no-FiLM baseline (dashed: thermal energy $k_\mathrm{B}T$ at $300$ K). \emph{Bottom:} offset-corrected energy vs. fine-tuned MACE-POLAR-1-M (per-charge mean removed, a metric generous to MACE-POLAR-1-M). E-MACE beats its MatPES backbone and matches MACE-POLAR-1-M on force, stays below $6$ meV/atom on raw energy RMSE, and beats MACE-POLAR-1-M on offset-corrected energy at every interpolation charge.}
\label{fig:emacevscharge}
\end{figure}

To test whether one set of weights covers the charge axis, we evaluate E-MACE on held-out VASP r$^2$SCAN sets at charges never seen in training: interpolation $q\in\{2e,4e,8e,12e,14e\}$ and extrapolation $q\in\{18e,20e\}$. Held-out geometries are drawn from AIMD at a nearby trained charge and relabeled at the target charge (Appendix~\ref{app:dataset}), so the test measures the vertical electronic force response to a change in charge at fixed, near-equilibrium nuclei, not full charge-plus-geometry generalization (the latter is exercised only by the MD runs). Figure~\ref{fig:emacevscharge} reports per-charge force and energy RMSE across all eleven charges. On the four training charges force RMSE is $5.5-7.5$ meV/\AA{} and per-atom energy RMSE below $0.15$ meV/atom. On the seven held-out charges force RMSE stays within $18-61$ meV/\AA{} and raw energy within $0.7-5.4$ meV/atom. Errors grow with $q$ and with the charge distance to the source trajectory that supplied the geometries. Among the sets relabeled $2e$ from their source ($q\in\{2e,8e,12e,18e\}$) force RMSE rises from $18$ to $44$ meV/\AA{}, the $4e$-relabeled sets ($q\in\{4e,14e,20e\}$) sit systematically higher, and there is no qualitative break between interpolation and extrapolation.

One set of E-MACE weights is therefore usable at any charge in the range without retraining. A single model, queried at the conditioning value of interest, replaces what would otherwise be a separate model per condition and provides a smooth interpolant between the labeled conditions. Cross-pipeline portability against an independent DFT engine (Appendix~\ref{app:crossdft}) indicates this reflects chemistry rather than artifacts specific to the DFT engine that produced the training labels.

Against the unconditioned fine-tuned MACE-MatPES backbone, E-MACE is about $3\times$ better on the training charges ($5.5-7.5$ vs. $16-28$ meV/\AA{}) and $1.3-2.8\times$ better at every held-out charge ($18-61$ vs. $45-97$ meV/\AA{}). Without a conditioning channel, the backbone cannot modulate its output with $q$, and E-MACE's advantage follows directly from the FiLM block supplying that mechanism.

A stronger comparison is fine-tuned MACE-POLAR-1-M \citep{batatia2026macepolar1}, a purpose-built charge-aware foundation about $20\times$ larger than E-MACE ($15.2$ M vs. $0.76$ M parameters) with explicit per-atom charge machinery. On forces the two are comparable across the whole range, held-out charges included. The lightweight scalar adapter matches the dedicated charge-aware architecture without supervising any internal quantity. Because the two use different absolute-energy references, the offset-corrected energies are compared (each charge's mean removed). This metric is generous to MACE-POLAR-1-M, since it scores only the within-charge spread and not the absolute per-charge baseline. Even so, E-MACE wins at every interpolation charge, by up to $\approx 16\times$ ($0.72$ vs. $11.8$ meV/atom at $q=12e$). The shape of the MACE-POLAR-1-M curve has a simple origin. Within either cell density of the dataset its energy spread is below $2$ meV/atom at every charge, but its energy baseline at fixed charge shifts between the two cell sizes by an amount growing roughly linearly with $q$, and a per-charge offset cannot remove a cell-size-dependent baseline. Its offset-corrected error therefore grows with $q$ wherever the geometry pool mixes both densities, and drops to nearly zero at $q\in\{16e,18e,20e\}$, whose pools derive from the single-density $q=16e$ trajectory (Appendix~\ref{app:dataset}). E-MACE stays below $0.8$ meV/atom on the same frames at every charge: the backbone reads the cell density from the geometry, so the conditioned model reconciles the two cell sizes.

\subsection{Data efficiency}

\begin{figure}[ht]
\centering
\includegraphics[width=0.8\linewidth]{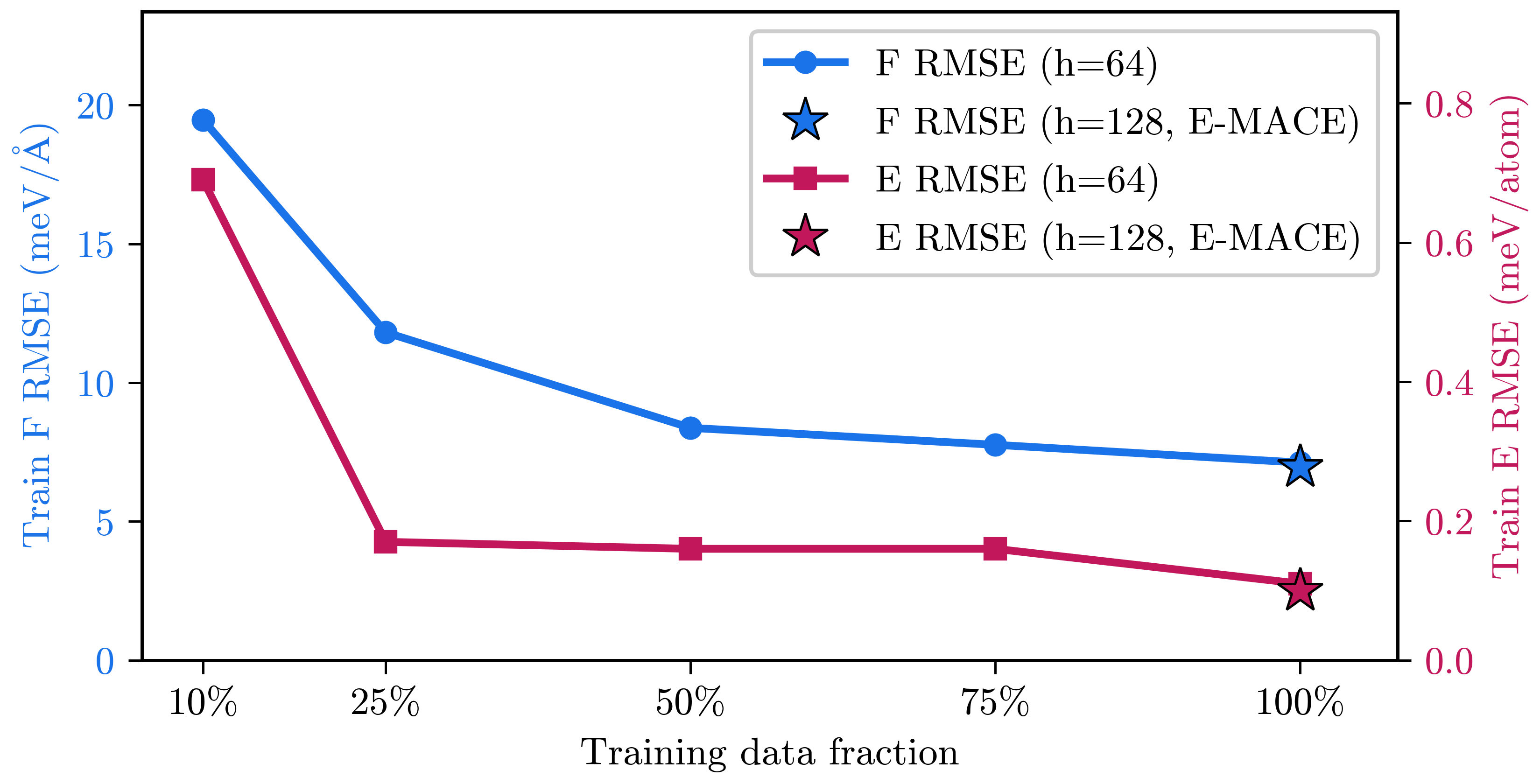}
\caption{Training charge mean force RMSE (left axis, blue) and energy RMSE (right axis, red) versus fraction of training data, for adapter widths $h=64$ (curves) and $h=128$ (stars at $100\%$). Energy saturates by $25\%$ training data. Force saturates by $50\%$.}
\label{fig:dataeff}
\end{figure}

Figure~\ref{fig:dataeff} reports the training charge force and energy RMSE of EquiFiLM applied to MACE-MatPES models trained on $10\%$, $25\%$, $50\%$, $75\%$ and $100\%$ of the combined training set. Energy RMSE saturates near $0.2$ meV/atom by the $25\%$ point (1,605 frames). Force RMSE saturates by the $50\%$ point (3,211 frames) at $\approx$$8$ meV/\AA{}. Doubling the adapter capacity to $h=128$ at $100\%$ data (the headline E-MACE configuration) drops energy RMSE further to $0.10$ meV/atom and force RMSE from $7.12$ to $6.96$ meV/\AA{}.

This sample efficiency reflects the inductive bias of the adapter: rather than re-discovering where the conditioning input should enter the network, EquiFiLM fixes its insertion point at every interaction layer and leaves only the per-channel gating values to be learned. A few thousand DFT-labeled frames spanning the conditioning axis are therefore sufficient to add a continuous external-conditioning channel to a foundation MLFF.

\subsection{Ablations}\label{sec:ablations}

To attribute E-MACE's improvement to specific design choices, a sequence of variants is trained on identical data and optimizer settings (Table~\ref{tab:ablations}): the no-FiLM baseline, a concat charge-embedding baseline that injects charge at the input rather than through per-layer gating, and, at fixed adapter width $h = 64$, an additive-only ablation that disables the multiplicative gate $\gamma$ (keeping only the shift $\beta$) and a random-initialization variant. A width comparison sets $h = 64$ against the headline $h = 128$. Inference cost is measured in a single back-to-back session on $324$-atom water cells on one A100 GPU in float32, with $90$ timed trials per variant.

The additive shift $\beta$ alone captures most of the gain. At $h = 64$, disabling $\gamma$ and keeping only $\beta$ cuts force RMSE from $21.3$ meV/\AA{} (no-FiLM) to $6.20$, comparable to the full $h = 64$ adapter ($7.12$ meV/\AA{}), and energy RMSE from $6.08$ to $0.09$ meV/atom. This fits a per-atom energy response that is roughly uniform across atoms at fixed $q$: each scalar feature picks up a charge-dependent offset that the readout composes with the unmodulated higher-rank channels to produce realistic forces. The multiplicative gate $\gamma$ would matter more for conditioning that rescales features rather than offsets them, such as temperature or pressure. It is kept in the default recipe because it is essentially free at inference and extends the same architecture to those axes.

Injecting the charge as an input token is not enough. The concat baseline concatenates a $16$-dimensional categorical charge embedding into the node features once at the input, in the spirit of SpookyNet-style conditioning. It reaches $20.8$ meV/\AA{} force and $6.08$ meV/atom energy RMSE, statistically indistinguishable from the unconditioned no-FiLM baseline and about $3\times$ worse on force than any FiLM variant. Both models have access to $q$, so the gap is not access but use: per-layer modulation lets the network adjust its internal features with charge at every depth, whereas a single input token does not.

\begin{table}[ht]
    \centering
    \caption{EquiFiLM ablation summary: held-out force and energy RMSE on each model's own $10\%$ validation split (final SWA epoch), with per-step inference cost. The four FiLM variants cluster tightly, E-MACE ($h=128$) is the headline. Two non-FiLM baselines are shown for reference, no-FiLM and a concatenated charge embedding: both are $\approx 3\times$ worse on force and $\approx 60\times$ on energy than any FiLM variant, and concat matches no-FiLM.}
    \label{tab:ablations}
    \begin{tabular}{lccc}
         \toprule
         \textbf{Variant} & \textbf{F RMSE} & \textbf{E RMSE} & \textbf{Inference} \\
                          & (meV/\AA{})       & (meV/atom)      & (\textmu s/atom-step) \\
         \midrule
         No-FiLM baseline                                       & 21.3  & 6.08 & $191 \pm 1$ \\
         Concat charge embedding                                & 20.8  & 6.08 & $191 \pm 1$ \\
         FiLM, $\beta$-only ($\gamma$ disabled)                 & 6.20  & 0.09 & $193 \pm 2$  \\
         FiLM, random initialization                            & 6.29  & 0.10 & $193 \pm 3$  \\
         FiLM, $h = 64$                                         & 7.12  & 0.11 & $193 \pm 2$  \\
         FiLM, $h = 128$ (\textbf{E-MACE})            & 6.96  & 0.10 & $193 \pm 2$ \\
         \bottomrule
    \end{tabular}
\end{table}

Random-initializing the FiLM weights rather than zero-initializing (both $h = 64$) gives $6.29$ meV/\AA{} and $0.10$ meV/atom, matching the zero-initialized adapter, and the $h = 128$ headline ($6.96$, $0.10$) sits in the same range. Across the four FiLM models force RMSE spans only $6.2-7.1$ meV/\AA{}, far below the $\approx 3\times$ gap to the no-FiLM baseline, so the gain is robust to adapter width, gating and initialization. Zero-initialization is kept because it makes E-MACE a drop-in replacement for the backbone before any gradient update, though it is not required for accuracy. All variants sit within $3\%$ of one another in per-step inference cost, so the FiLM head adds no measurable overhead.

\subsection{Applications}

E-MACE is now applied to molecular dynamics (MD) simulations across the full charge axis, with one set of weights driving trajectories at charges spanning training, interpolation and extrapolation. All MD experiments in the main text use the $324$-atom training cell (cubic, edge $14.8$ \AA{}) at $300$ K with a $0.5$ fs timestep, and the headline E-MACE configuration.

\subsubsection{Energy conservation}

MD integrates the equations of motion under the model's forces. If those forces are not the gradient of a consistent energy, as can happen when a conditioning channel introduces non-conservative pathways, the total energy drifts and the trajectory fails to sample the correct distribution. The standard test equilibrates in the canonical (NVT) ensemble and then detaches the thermostat to continue in the microcanonical (NVE) ensemble, where total energy must be conserved by construction.

\begin{figure}[ht]
\centering
\includegraphics[width=0.8\linewidth]{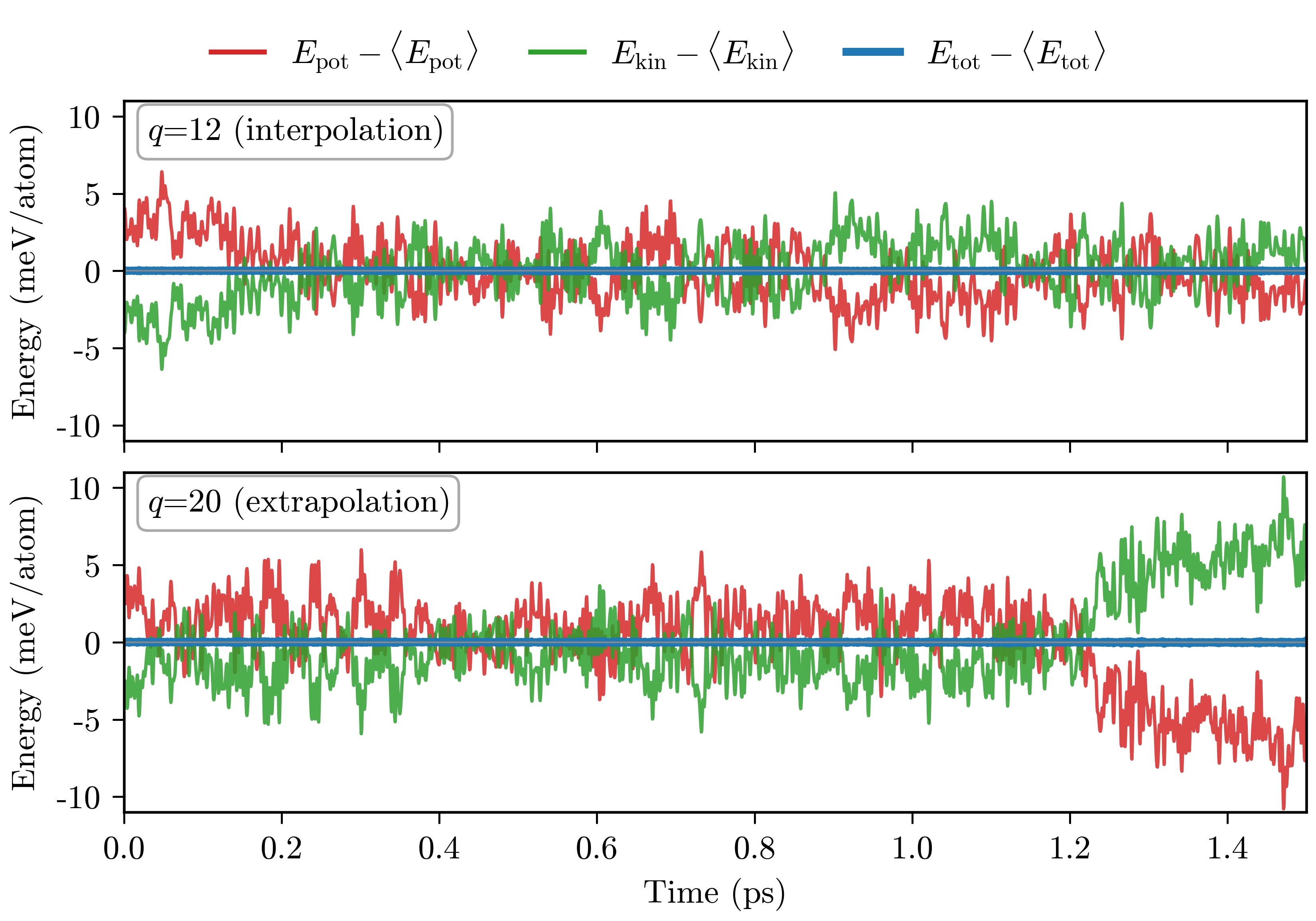}
\caption{Energy conservation over $1.5$ ps post-thermostat detach at $q=12e$ (interpolation, top) and $q=20e$ (extrapolation, bottom). $E_{\rm pot}$ (red) and $E_{\rm kin}$ (green) exchange thermal-scale energy of order $\pm 5$ meV/atom while their sum $E_{\rm tot}$ (blue) stays within an $\approx 0.18$ meV/atom band, with mean drift below $0.01$ meV/atom. One set of E-MACE weights gives stable dynamics across the full interpolation-extrapolation range.}
\label{fig:md_stability}
\end{figure}

Over a $1.5$ ps post-thermostat detach NVE window the mean total-energy drift is $+0.004$ meV/atom at $q=12e$ (interpolation) and $+0.010$ meV/atom at $q=20e$ (extrapolation), shown in Figure~\ref{fig:md_stability}. Because the FiLM block depends only on the per-graph scalar and acts only on scalar channels, the conditioning channel preserves the property that forces are exact gradients of a position-dependent energy uniformly across $q$. The full NVT-NVE protocol is described in Appendix~\ref{app:nveconservation}.

\subsubsection{Charge-dependent diffraction signal}

X-ray and electron scattering probe the reduced pair distribution function (PDF) $G(r) = 4\pi r \rho_0 [g_{\rm tot}(r) - 1]$, where $g_{\rm tot}$ sums the O-O, O-H and H-H partials weighted by their squared atomic scattering factors (X-ray weights $0.64/0.32/0.04$, so the O-O term dominates). Ultrafast electron diffraction (UED) of photoexcited water measures the time-resolved difference $\Delta G(r;t)$ following the perturbation \citep{lin2021ued}. A model that runs stable MD across the charge axis predicts the corresponding equilibrium charge response $\Delta G(r;q) = G(r;q) - G(r;0)$ at any $q$.

\begin{figure}[ht]
\centering
\includegraphics[width=0.8\linewidth]{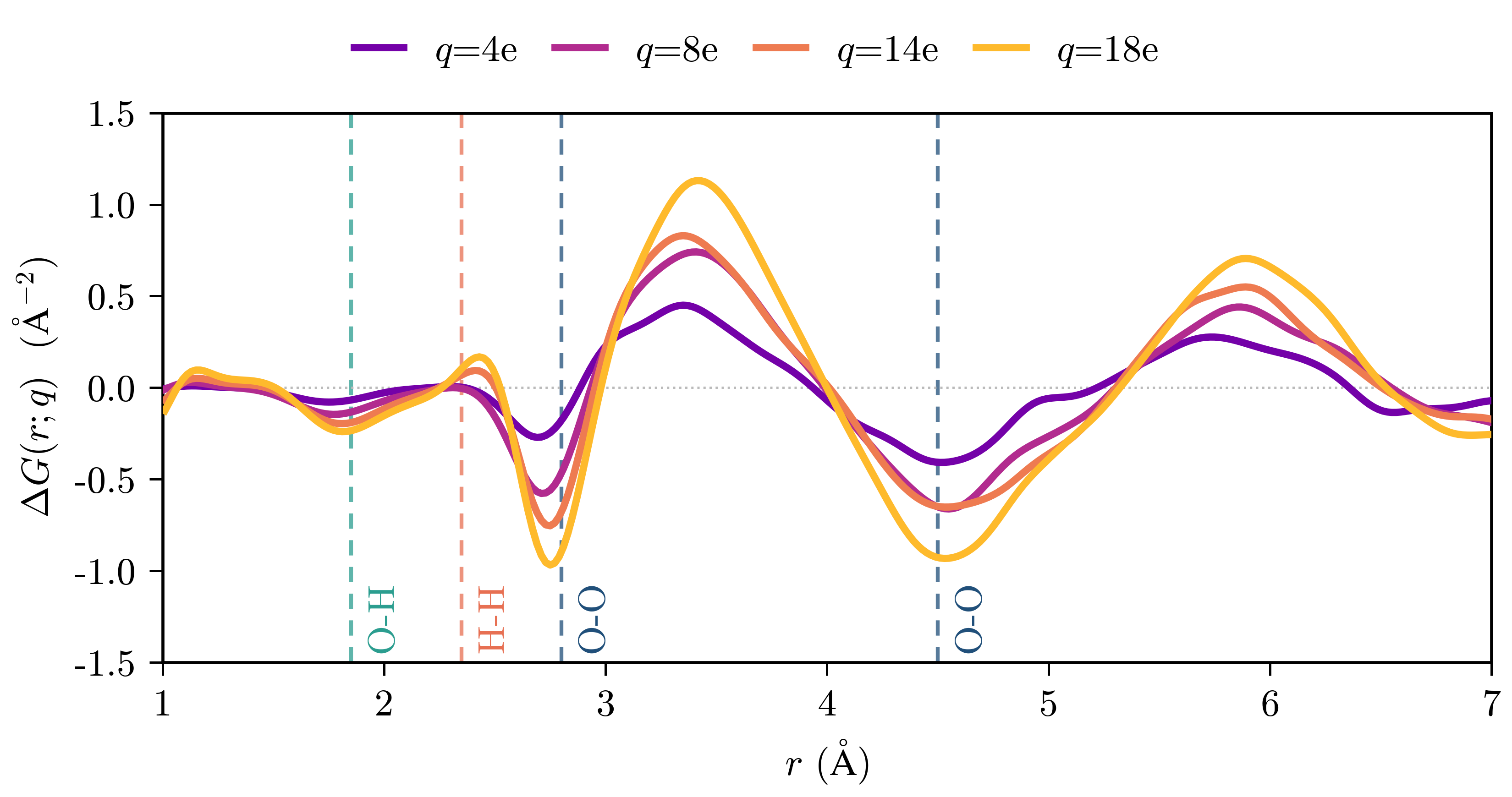}
\caption{Equilibrium charge response of the reduced PDF, $\Delta G(r;q)=G(r;q)-G(r;0)$, from E-MACE NVT trajectories at $q\in\{4e,8e,14e,18e\}$, computed with X-ray scattering weights. Dashed vertical lines mark the ambient-water O-H, H-H and O-O pair distances. This is the equilibrium charge-response component of the reduced PDF difference that ultrafast electron diffraction and X-ray pair-distribution-function analysis measure.}
\label{fig:delta_pdf}
\end{figure}

Figure~\ref{fig:delta_pdf} shows $\Delta G(r;q)$ from NVT trajectories at three interpolation charges ($q\in\{4e,8e,14e\}$) and one extrapolation charge ($q=18e$). A trough near $r\approx 2.7$ \AA{} and a peak near $r\approx 3.3$ \AA{} mark an outward shift of the first O-O coordination shell. Both grow monotonically with $|q|$, and the $q=18e$ curve continues the trend without discontinuity. This is a concrete and falsifiable structural prediction for water under ionization, and one model supplies it at any $q$ within picoseconds of MD per query. At the highest charges the H-H partial also grows a peak at the H$_2$ bond length of $\approx 0.74$ \AA{}, the signature of reductive H$_2$ formation (H-H panel of Figure~\ref{fig:gr_pairs} in Appendix~\ref{app:pdf}). This feature is already present in the $q=16e$ AIMD training data, so the model reproduces a genuine reductive process rather than a structural artifact. Appendix~\ref{app:pdf} also gives the per-pair correlations $g_{\alpha\beta}(r;q)$ and the smoothing convention.

All results above use the $324$-atom training cell. In a $2\times 2\times 2$ supercell of $2,592$ atoms, eight times larger than any configuration the model was trained on, the same set of weights continues to run stable, energy-conserving MD across the charge range and reproduces both the per-pair structural responses and $\Delta G(r;q)$ quantitatively (Appendix~\ref{app:supercell}). The charge-conditioned dynamics therefore transfer beyond the training cell size as well as beyond the training charges.

\section{Discussion}

Charged liquid water is one instance of a capability that is general by construction. The adapter is a small per-graph MLP whose output gates the scalar slice of every interaction layer, nothing in this mechanism refers to charge. Because the gating depends only on a generic continuous scalar and acts only on E(3)-invariant channels, replacing total charge with another smooth external variable changes the input the gating MLP reads, not the architecture. Natural targets include variable temperature dynamics under one set of weights, field-response coefficients computed as energy derivatives, hydrostatic pressure conditioning, and doping fraction interpolation. In principle, each needs only a fine-tuning corpus spanning the relevant axis. 

The construction is equally independent of the backbone. Any equivariant message-passing potential with scalar interaction-layer channels can host the block, including the MACE, NequIP \citep{batzner2022nequip}, Allegro \citep{musaelian2023allegro} and Equiformer families and foundations such as UMA, eSEN and SevenNet-Omni, because the adapter never touches the higher-rank geometric channels that distinguish these architectures. Training a charge-aware foundation from scratch needs corpora of order $10^8$ structures. EquiFiLM adds a conditioning axis for a few thousand DFT-labeled frames, provided the foundation already covers the chemistry. The broader point is that conditioning a foundation MLFF is an adapter problem, just as task adaptation became one in language and vision. Parameter-efficient modulation of the forward pass brings a new controllable degree of freedom under the model without retraining the representation. Wherever the backbone already covers the chemistry and a small corpus spans the axis, as in photoexcitation, electrochemical interfaces, plasma-driven solvation or doped semiconductors, the conditioning axis can be added at adapter cost rather than foundation cost.

EquiFiLM presupposes that the foundation backbone already covers the chemistry of interest at near-DFT accuracy on the unconditioned baseline. Chemistry outside that coverage reduces the recipe to standard fine-tuning, where most of the work falls on the backbone weights rather than on the adapter. The conditioning effect is also assumed to be smooth in $c$ over the range of interest, and the relative loading on $\beta$ versus $\gamma$ depends on whether the conditioning acts predominantly as a baseline shift (as for charge, Section~\ref{sec:ablations}) or as a feature rescaling, the latter of which has not been validated empirically. Beyond a single continuous scalar, conditioning on a discrete or categorical axis would require a different input embedding, and multi-axis conditioning is untested, with no guarantee that the $\beta$ and $\gamma$ loading factorizes across axes.

\section{Related work}

EquiFiLM is a Feature-wise Linear Modulation operator \citep{perez2018film} applied to the scalar channels of an equivariant network. FiLM generalizes Conditional Batch Normalization \citep{devries2017cbn}, where the parameters of normalization layers are produced by an auxiliary network conditioned on a side input. The broader parameter-efficient fine-tuning literature, including LoRA \citep{hu2022lora}, Houlsby Adapters \citep{houlsby2019adapters} and HyperNetworks \citep{ha2017hypernetworks}, establishes the same framing in language and vision: small trainable modules added to a frozen or jointly fine-tuned backbone, with footprints orders of magnitude below full retraining. EquiFiLM is in this terminology a conditional adapter for an equivariant atomistic network, in which the gating depends on a per-input continuous scalar rather than being merged into the base weights at inference (as in LoRA) or being task-fixed (as in Houlsby Adapters). To our knowledge it is the first such conditional adapter for an equivariant interatomic potential.

Charge-aware atomistic networks prior to ours fall into two families. SpookyNet \citep{unke2021spookynet} injects total charge and spin as additive embeddings at the input layer of an attention-based message-passing architecture. This differs operationally from per-layer modulation of a pretrained backbone, and SpookyNet is trained from scratch rather than as an adapter on top of a foundation. The charge-equilibration and Wannier-centroid family \citep{batatia2026macepolar1,ko20214ghdnnp,kalita2025aimnet2nse,zhang2022dplr,zubatyuk2021aimnetnse,calegariandrade2023,gao2024dpmp,gao2025polarizablefoundation} predicts per-atom partial charges or Wannier-centroid positions and feeds them back through an explicit Coulomb sum, sometimes with iterative equilibration, trained from scratch with dedicated charge supervision. EquiFiLM commits only to the existence of scalar channels at every interaction layer, treats the conditioning input as a generic continuous variable, and supervises no internal physical quantity. As a direct empirical comparison, fine-tuning the charge-aware foundation MACE-POLAR-1-M \citep{batatia2026macepolar1} on the same data yields force accuracy competitive with E-MACE on the training charges, despite its explicit per-atom charge machinery and dedicated charge supervision. The lightweight adapter matches a dedicated charge-aware foundation on the training charges without supervising any internal quantity. The two approaches are not mutually exclusive: an EquiFiLM adapter could in principle be combined with an explicit charge-aware mechanism.

\section{Conclusions}

EquiFiLM extends the parameter-efficient adapter pattern to equivariant foundation interatomic potentials. It is demonstrated on charged liquid water with MACE-MatPES as the backbone, calling the resulting model E-MACE. The same recipe applies, without architectural change, to any equivariant MLFF with scalar interaction-layer channels and to any continuous external variable on which the energy depends smoothly. The practical implication for machine learning in materials science is that adding a continuous control variable to a foundation MLFF, whether charge or temperature, is the same adapter problem that language and vision already solved: a small fine-tuning corpus along the conditioning axis is enough, without retraining the foundation.

\section*{Data and Code Availability}
The reference implementation, including the four MACE source-file patches that introduce \texttt{ChargeFiLMBlock}, training and evaluation example scripts, unit tests, and a reproduction walkthrough, is available at \href{https://github.com/samsahsch/EquiFiLM}{\texttt{https://github.com/samsahsch/EquiFiLM}}. The training data (four AIMD trajectories at $q\in\{0,6e,10e,16e\}$, each at the r$^2$SCAN meta-GGA level) and the trained \texttt{E-MACE} checkpoint are deposited at Zenodo DOI:
\href{https://zenodo.org/records/20067517}{10.5281/zenodo.20067517}.

\begin{ack}
 This work was performed using National Energy Research Scientific Computing Center (NERSC), a Department of Energy User Facility under NERSC award BES-ERCAP 0035639. This work was supported by the AMOS program within the U.S. Department of Energy, Office of Science, Basic Energy Sciences, Chemical Sciences, Geosciences, and Biosciences Division. S.S.S. and T.L. acknowledge support by the U.S. Department of Energy, Office of Science, Office of Basic Energy Sciences under Contract No. DE-AC02-76SF00515. A.N. was supported by Office of Naval Research through a Multi-University Research Initiative (MURI) grant N00014-24-1-2313.
\end{ack}

{
\bibliographystyle{unsrtnat}
\bibliography{refs}
}
\clearpage


\appendix

\centerline{\LARGE\textbf{Appendix}}

\section{Per-state cross-evaluation matrix}
\label{app:crossmatrix}

The introduction claims that per-state MACE-MatPES specialists trained one model per charge cannot generalize to unseen charges. This appendix provides the supporting cross-evaluation data, comparing four specialists (each fine-tuned on a single training charge $q\in\{0,6e,10e,16e\}$) against the foundation backbone, the unconditioned no-FiLM MACE-MatPES baseline, and E-MACE. Every model is evaluated on the same fixed validation set: a frozen $10\%$ random hold-out of every training trajectory (Appendix~\ref{app:dataset}). Each per-state specialist is fine-tuned from the MACE-MatPES backbone with the same optimizer settings and total training budget as E-MACE (Appendix~\ref{app:hparams}). The only difference between specialists and E-MACE is the data their gradient updates see and the absence of the FiLM adapter.

\begin{figure}[!ht]
\centering
\includegraphics[width=0.8\linewidth]{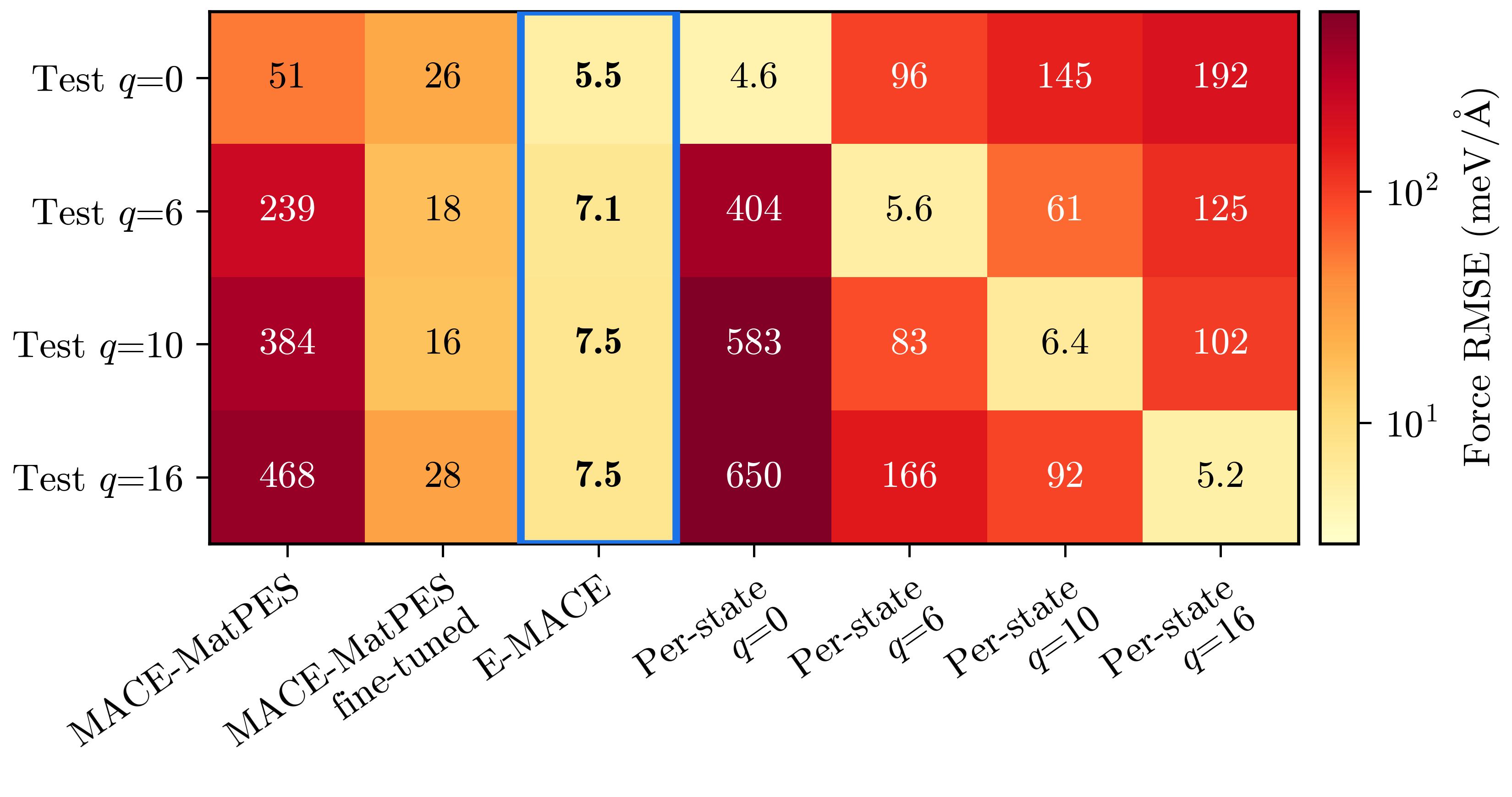}
\includegraphics[width=0.8\linewidth]{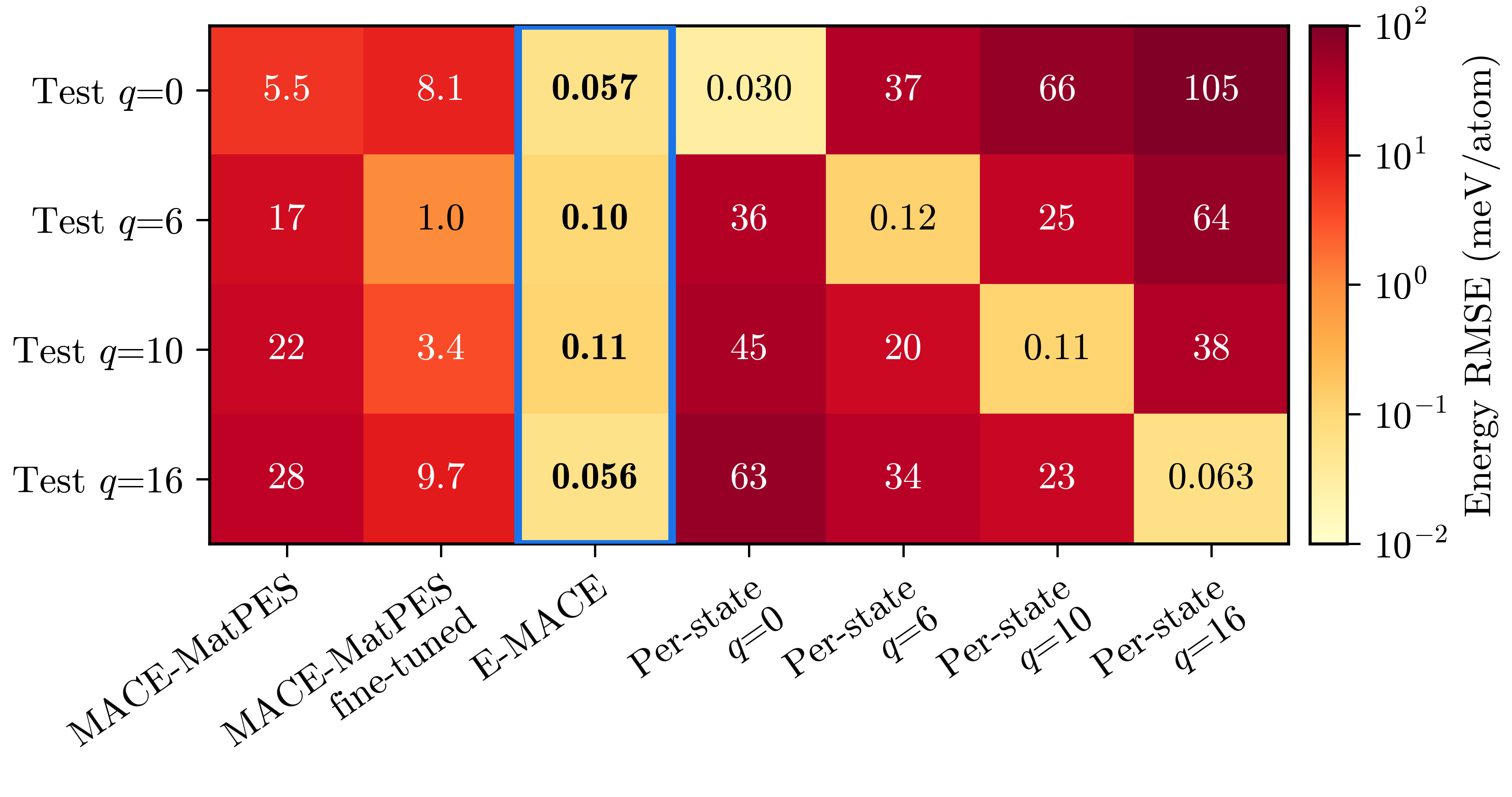}
\caption{Cross-evaluation heatmaps on the four training charges. \emph{Top:} force RMSE in meV/\AA{}. \emph{Bottom:} energy RMSE in meV/atom. Rows are the test charge, columns the model. The four right-most columns are per-state MACE-MatPES specialists, each fine-tuned on a single training charge. The bright-yellow diagonal in those columns marks each specialist's matched training charge. The leftmost column is the unconditioned MACE-MatPES foundation backbone (zero-shot, no fine-tuning). The second column is the no-FiLM baseline (single MACE-MatPES fine-tuned on all four training charges combined, no FiLM block). The third column is E-MACE (same backbone, same data, plus the FiLM adapter).}
\label{fig:figS_heatmap}
\end{figure}

Figure~\ref{fig:figS_heatmap} shows the resulting force and energy heatmaps, and Table~\ref{tab:crossmatrix} lists the per-state specialist cells numerically. Each specialist is excellent on its own charge and collapses everywhere else. Relative to the specialist matched to the test charge, off-diagonal cells degrade by roughly $200$ to $3,500\times$ in energy and $10$ to $130\times$ in force. The off-diagonal magnitude grows monotonically in $|q_{\rm trained}-q_{\rm test}|$: the smallest off-diagonal cells are the nearest-charge pairs (e.g.\ $q=6e$ specialist evaluated at $q=10e$ at $82.5$ meV/\AA{} in force), and the largest are the most-separated pairs (e.g.\ $q=0$ specialist evaluated at $q=16e$ at $650$ meV/\AA{}). Within a fixed charge gap the matrix shows mild direction-dependent variation at the level of a factor of two to three, and the direction is not consistent across metrics ($q=0$ specialist on $q=16e$ gives $650$ meV/\AA{} in force while $q=16e$ specialist on $q=0$ gives $192$, but on energy the asymmetry runs the other way at $62.8$ versus $104.6$ meV/atom). The off-diagonal degradation is therefore set primarily by the charge gap rather than by the direction of extrapolation. The empirical content of this matrix is that a model trained on a single charge has no representation of how the potential energy surface depends on the conditioning input: at every off-diagonal cell, regardless of direction, accuracy drops by orders of magnitude. This is the observation that motivates a charge-conditioned model.

\begin{table}[ht]
\centering
\caption{Per-state specialist cross-evaluation. Each cell reports \textit{energy RMSE (meV/atom) / force RMSE (meV/\AA{})}. Rows: the charge the specialist was fine-tuned on. Columns: test charge.}
\begin{tabular}{lcccc}
\toprule
\multirow{2}{*}{Specialist trained on}
   & \multicolumn{4}{c}{Test charge}  \\
\cmidrule(lr){2-5}
   & $q=0$ & $q=6e$ & $q=10e$ & $q=16e$ \\
\midrule
$q=0$    & $0.030$ / $4.6$  & $36.2$ / $404$  & $44.8$ / $583$  & $62.8$ / $650$ \\
$q=6e$   & $36.9$ / $96.1$  & $0.12$ / $5.6$  & $20.0$ / $82.5$ & $33.9$ / $166$ \\
$q=10e$  & $66.4$ / $145$   & $25.4$ / $60.5$ & $0.11$ / $6.4$  & $22.5$ / $92.4$ \\
$q=16e$  & $104.6$ / $192$  & $63.7$ / $125$  & $38.0$ / $102$  & $0.063$ / $5.2$ \\
\bottomrule
\end{tabular}
\label{tab:crossmatrix}
\end{table}

The three reference columns of Figure~\ref{fig:figS_heatmap} show why conditioning, rather than more training data, is the missing ingredient. The zero-shot foundation backbone is uniformly poor and degrades further at high $q$: it never specializes, so it fails everywhere. The no-FiLM baseline shows the complementary failure. Trained on all four charges but given no input that distinguishes them, it can only learn a single compromise potential energy surface, so its errors are roughly uniform across charges ($16$ to $28$ meV/\AA{} in force) yet remain far above every specialist diagonal. E-MACE removes the compromise. With charge as a conditioning input, one set of weights reaches $5.5$ to $7.5$ meV/\AA{} across the four charges, within $1$ to $2$ meV/\AA{} of each specialist on its own diagonal ($4.6$ to $6.4$ meV/\AA{}), and matches the specialist diagonals outright on energy ($0.056$ to $0.11$ vs. $0.030$ to $0.12$ meV/atom, Table~\ref{tab:crossmatrix} and Figure~\ref{fig:figS_heatmap}). One conditioned model therefore delivers near-specialist accuracy at every charge simultaneously, which no unconditioned model in the matrix achieves at any charge. This is the observation that motivates a charge-conditioned model.

\section{FiLM adapter parameter and inference cost}
\label{app:filmcost}

Each ChargeFiLMBlock contains two independent two-layer MLPs (one for $\gamma$, one for $\beta$), each mapping $\mathbb{R} \to \mathbb{R}^{N_s}$ with hidden width $h=128$ and SiLU activation. One \mbox{ChargeFiLMBlock} is attached per interaction layer, at $K=2$ sites in the MACE-MatPES network (\texttt{interactions.0} and \texttt{interactions.1}).

The two interaction layers differ in their scalar channel count due to the MACE-MatPES hidden irreps: $N_s = 128$ for layer~0 and $N_s = 256$ for layer~1. The per-network parameter count is $(1 \cdot h + h) + (h N_s + N_s)$ (first linear layer: $1 \to h$ weights+bias; second: $h \to N_s$ weights+bias), giving $16,768$ per network for layer~0 ($N_s = 128$) and $33,280$ for layer~1 ($N_s = 256$). Each block has two networks ($\gamma$ and $\beta$), so block-level counts are $33,536$ and $66,560$ respectively. The total adapter footprint is therefore $33,536 + 66,560 = 100,096$ parameters ($\approx 0.10$ M). Inspecting the trained checkpoint confirms the MACE-MatPES backbone contributes $655,534$ parameters ($\approx 0.66$ M), making the adapter a $\approx 15\%$ overhead (total model $\approx 0.76$ M).

The benchmark times one forward-and-backward pass of each model on a fixed $324$-atom liquid water cell at batch size $1$ on a single A100 GPU, timing $90$ trials after $10$ warm-up trials. A forward-and-backward pass is what an MD step requires, since forces are obtained as the gradient of the energy with respect to atomic positions. Two protocol details matter for a fair comparison. First, every model is timed in the same dtype, \texttt{float32}, the precision used for production MD. Timing in \texttt{float64} inflates all numbers by roughly $1.3\times$ on A100. Second, each timed trial is run on a freshly perturbed copy of the geometry (atomic positions jittered by $10^{-3}$ \AA{}), because the ASE calculator interface caches results for an unchanged configuration and would otherwise return the previous step's forces without recomputation, yielding spuriously fast times. All five rows of Table~\ref{tab:inference_time} were measured in a single back-to-back session on one idle A100 under these settings, so they are directly comparable.

Under this protocol Table~\ref{tab:inference_time} compares five models. E-MACE measures $193 \pm 2$ \textmu s per atom-step against $191 \pm 1$ \textmu s for the unconditioned MACE-MatPES backbone fine-tuned on the same data. The two differ by $\approx 1\%$ with overlapping $1\sigma$ bars and are statistically indistinguishable, as both share the identical H/O backbone and differ only by the FiLM head. The per-step cost of the FiLM adapter itself is one MLP forward pass on a single scalar input plus $K$ element-wise operations on the scalar slices of the message tensor, an $O(N_s)$ contribution dominated at $N$-atom scale by the equivariant tensor products of the backbone. The remaining two rows are not the same architecture and are included for context. The zero-shot MACE-MatPES foundation measures $213 \pm 3$ \textmu s per atom-step. It is slower than the fine-tuned backbone not because of any adapter but because it is the full-periodic-table foundation ($9.06$ M parameters spanning $\approx 89$ elements) and carries that entire element-embedding and readout machinery through every forward pass even on a two-element water cell, whereas the fine-tuned backbone is the same architecture filtered to hydrogen and oxygen ($0.66$ M parameters) by the standard foundation fine-tuning workflow (Appendix~\ref{app:hparams}). MACE-POLAR-1-M and its fine-tune measure $\approx 578-582$ \textmu s per atom-step, about $3\times$ E-MACE, reflecting the larger purpose-built charge-aware architecture rather than any difference in conditioning mechanism. The two MACE-POLAR-1-M rows are equal within error, as expected since fine-tuning does not change the architecture.

NequIP and Allegro are omitted from Table~\ref{tab:inference_time} because their reference inference uses a different framework and a compiled, CUDA-only kernel stack that is not directly comparable to the eager-mode PyTorch timing of the MACE family reported here. Measured under the same conditions, they run faster than the MACE family ($\approx 27$ and $\approx 32$ $\mu$s per atom-step, respectively), but this reflects their smaller architectures and a different compiled runtime rather than anything about the conditioning mechanism.

The inference column of Table~\ref{tab:ablations} extends the same measurement to the FiLM ablation variants. $\beta$-only, random initialization, $h = 64$ and $h = 128$ all measure $193$ \textmu s per atom-step, inside the same $191-193$ \textmu s band, so neither adapter width, gating, nor initialization changes the per-step cost.

A naive PyTorch implementation of the FiLM scalar-slice update via in-place index assignment, \texttt{out = features.clone(); out[:, :N\_s] = s\_film}, where \texttt{s\_film} denotes the FiLM-transformed scalar slice of Equation~\ref{eq:filmblock}, triggers the slow IndexPut backward path during force computation, which routes through a generic strided-write kernel rather than the contiguous concatenation operator. Because forces require a second backward pass through the model (gradient of energy with respect to positions), at $324$-atom scale this routing slows a forward-and-backward step by roughly two orders of magnitude. The reference implementation instead builds the output tensor in pure-functional form, \texttt{out = torch.cat([s\_film, rest], dim=-1)}, which restores native message-passing speed and is mathematically identical to the in-place version. Without this rewrite E-MACE would be approximately $100\times$ slower than its own backbone.

\section{Dataset composition}
\label{app:dataset}

This appendix specifies the four AIMD trajectories used as training labels, the DFT settings used to compute per-frame energies and forces, the protocol used to sample held-out geometries at each of the seven evaluation charges, and the fixed train/validation split.

The training set is the union of four ab initio molecular dynamics (AIMD) trajectories of liquid water at total electronic charges $q\in\{0,6e,10e,16e\}$, where $q$ counts the number of electrons added to a closed-shell $324$-atom neutral reference cell ($108$ H$_2$O molecules under periodic boundary conditions). Trajectories are propagated under VASP \citep{kresse1996vasp} at the r$^2$SCAN meta-GGA level \citep{furness2020r2scan} at $300$ K, with energies and forces written at every step. Per-state frame counts and the distribution across two cubic cell edge lengths ($13.8$ \AA{} and $14.8$ \AA{}) are listed in Table~\ref{tab:dataset}. Of the $6,423$ total frames, $1,500$ are at the dense edge length and $4,923$ at the sparse edge length, an artifact of the AIMD setup ($q=16e$ contains no $13.8$ \AA{} segment).

\begin{table}[ht]
\centering
\caption{Per-state composition of the training AIMD trajectories. Each frame is a $324$-atom snapshot ($108$ H$_2$O) under periodic boundary conditions.}
\begin{tabular}{lcccc}
\toprule
State & Frames & At $13.8$ \AA{} (dense) & At $14.8$ \AA{} (sparse) & Dense / sparse \\
\midrule
$q=0$    & $1,455$ & $500$ & $955$    & $34/66\%$ \\
$q=6e$   & $2,000$ & $500$ & $1,500$ & $25/75\%$ \\
$q=10e$  & $2,000$ & $500$ & $1,500$ & $25/75\%$ \\
$q=16e$  & $968$    & $0$   & $968$    & $0/100\%$ \\
\midrule
\textbf{Total} & $\mathbf{6,423}$ & $\mathbf{1,500}$ & $\mathbf{4,923}$ & $\mathbf{23/77\%}$ \\
\bottomrule
\end{tabular}
\label{tab:dataset}
\end{table}

Energies and forces at every snapshot, both during the training trajectory AIMD and during the held-out single-point evaluations described below, are computed self-consistently in VASP at the r$^2$SCAN meta-GGA level under the settings of Table~\ref{tab:vasp_settings}. The total electron count is set as $\texttt{NELECT} = N_{\rm neutral} + q$, where $N_{\rm neutral}$ is the closed-shell valence-electron count of the neutral $324$-atom reference under the chosen PAW pseudopotentials. By convention $q\geq 0$ denotes electrons added to the cell, populating the lowest unoccupied bands of the neutral reference. For scale, the largest training charge $q=16e$ corresponds to $0.049$ electrons per atom, one excess electron per roughly seven water molecules, or $1.9\%$ of the valence electrons in the cell. This jellium-compensated setup represents the delocalized, high-density excess electron limit relevant to ultrafast photoinjection and electron-transfer experiments. It does not, and a uniform neutralizing background cannot, represent a localized solvated-electron polaron, so we use ``excess electron'' rather than ``hydrated electron'' language and restrict the plasma-solvation analogy to the delocalized regime.

All training and held-out evaluation charges are even, which makes every cell closed-shell at $\texttt{ISPIN}=1$. Odd-$q$ configurations would require open-shell treatment and are out of scope. In a periodic cell the added electrons are compensated by a uniform neutralizing background (the standard convention for charged supercells), so the excess charge is delocalized across the cell rather than localized as a solvated electron with an explicit counter-cation. The conditioning variable $q$ therefore controls a homogeneous excess electron density regime. Modeling localized charge carriers or explicit counterions would require a different reference and is outside the present scope. A charged periodic cell also carries a spurious electrostatic self-energy between the net charge and its neutralizing background and periodic images, scaling as $q^2/L$ at fixed cell size. We apply no Makov-Payne-type finite-size correction because this term is a per-charge constant that enters the DFT label and the model prediction identically, and is absorbed into the per-charge baseline that $\beta(q)$ learns. Forces are insensitive to the uniform background, which is why the force-based results are the physically robust ones. Two numerical settings warrant a caveat at high charge. The $\Gamma$-only $k$-sampling under-resolves the band dispersion of the added delocalized states, and the $0.05$ eV occupation smearing ($\approx 580$ K electronic temperature) sets the partial filling of the LUMO manifold. Both are held fixed across the training and held-out single-point evaluations, so they do not bias the per-charge comparison, but absolute high-$q$ forces carry a corresponding convergence uncertainty.

\begin{table}[ht]
\centering
\caption{Shared VASP r$^2$SCAN settings. All values apply identically to the AIMD training snapshots and to the held-out per-charge single-point evaluations.}
\begin{tabular}{ll}
\toprule
Parameter & Value \\
\midrule
Functional & r$^2$SCAN meta-GGA (\texttt{METAGGA = R2SCAN}) \\
Plane-wave cutoff & $550$ eV (\texttt{ENCUT}) \\
$k$-point sampling & $\Gamma$-point only ($1\times 1\times 1$) \\
PAW pseudopotentials & VASP PAW for O and H (PBE-derived, used with r$^2$SCAN) \\
Smearing scheme & Gaussian (\texttt{ISMEAR}=0) \\
Smearing width & $0.05$ eV at $q\neq 0$; $0.001$ eV at $q=0$ \\
SCF tolerance & $10^{-6}$ eV (\texttt{EDIFF}) \\
Spin treatment & non-spin-polarized (\texttt{ISPIN}=1) at all $q$ \\
Total electron count & $\texttt{NELECT} = N_{\rm neutral} + q$ \\
\bottomrule
\end{tabular}
\label{tab:vasp_settings}
\end{table}

For the seven held-out charges no AIMD trajectory is available. For each held-out $q$, we sample geometries evenly across the AIMD trajectory at a nearby training charge and re-evaluate them self-consistently under the same VASP r$^2$SCAN protocol at the new \texttt{NELECT}. The per-target-charge source trajectories are listed in Table~\ref{tab:heldout_sampling}. This protocol mirrors standard practice for generating per-charge DFT labels in condensed-phase systems where a full AIMD trajectory at every target charge is computationally infeasible. Equilibrium geometries at the source charge serve as a representative pool of nuclear configurations, and the model is asked to predict the change in forces and per-atom energy that results from perturbing the electronic state without first re-equilibrating the nuclei.

\begin{table}[ht]
\centering
\caption{Source AIMD trajectory used to sample geometries for each held-out evaluation charge.}
\begin{tabular}{lcc}
\toprule
Held-out $q$ & Category & Source AIMD trajectory \\
\midrule
$2e$  & interpolation  & $q=0$       \\
$4e$  & interpolation  & $q=0$       \\
$8e$  & interpolation  & $q=6e$      \\
$12e$ & interpolation  & $q=10e$     \\
$14e$ & interpolation  & $q=10e$     \\
$18e$ & extrapolation  & $q=16e$     \\
$20e$ & extrapolation  & $q=16e$     \\
\bottomrule
\end{tabular}
\label{tab:heldout_sampling}
\end{table}

For every model variant we use a uniform $10\%$ random hold-out within each training charge AIMD trajectory, fixed across variants by random seed. This split provides the validation set during SWA and is the source of every training charge accuracy number reported in the body and in Appendix~\ref{app:crossmatrix}. The held-out-charge sets combine geometries drawn evenly from a source trajectory with a charge the model never trained on, so every geometry-charge combination they contain is new to the model. Evaluating familiar geometries at a new charge is what isolates the vertical electronic response of Section~3.1. Generalization to new geometries and new charges at once is probed by the MD simulations.

\section{Hyperparameters for all trained variants}
\label{app:hparams}

All MACE-based variants (E-MACE, the no-FiLM baseline, and the four FiLM ablations of Section~\ref{sec:ablations}) share the training settings listed in Table~\ref{tab:shared_hparams}. The loss is the standard MACE per-atom weighted sum of energy and force terms with default weights ($w_E = 1$, $w_F = 100$). Stochastic weight averaging \citep{izmailov2018swa} starts at epoch $150$ and runs to $250$ epochs for every variant. The SWA-averaged weights at the final epoch are exported as the model.

\begin{table}[ht]
\centering
\caption{Settings shared across all MACE-based training runs. Hardware is a single NVIDIA A100 $40$ GB GPU. Per-variant training takes approximately $67$ GPU-hours.}
\begin{tabular}{ll}
\toprule
Setting & Value \\
\midrule
Foundation backbone    & MACE-MatPES \citep{kaplan2025MatPES} (architecture, initial weights, $E_0$s) \\
Optimizer              & AdamW \citep{loshchilov2019adamw} with AMSGrad \citep{reddi2018amsgrad} \\
Learning rate          & $0.01$ \\
Batch size             & $2$ \\
Weight EMA decay       & $0.99$ \\
Float dtype            & \texttt{float64} \\
Edge cutoff $r_{\max}$ & $6.0$ \AA{} \\
Data scaling           & \texttt{rms\_forces\_scaling} \\
Loss weights           & $w_E = 1$, $w_F = 100$ \\
Validation split       & uniform $10\%$ random hold-out per training trajectory \\
SWA schedule           & \texttt{start\_swa=150}, \texttt{max\_num\_epochs=250} \\
\bottomrule
\end{tabular}
\label{tab:shared_hparams}
\end{table}

All variants are fine-tuned through MACE's foundation-model workflow with element filtering (\texttt{--foundation\_filter\_elements}). The model inherits the MACE-MatPES architecture from the released \texttt{MACE-MatPES-r2scan-omat-ft} checkpoint (\texttt{hidden\_irreps}\,$=$\,\texttt{128x0e+128x1o}, two interaction layers, correlation order $3$, $r_{\max}=6$ \AA{}) and its isolated-atom reference energies, but restricts the element set to the hydrogen and oxygen present in water. This reduces the parameter count from the full-periodic-table foundation ($9.06$ M parameters spanning $\approx 89$ elements) to $0.66$ M for the H/O backbone, onto which the $0.10$ M EquiFiLM adapter is added for the FiLM variants (Appendix~\ref{app:filmcost}). The filtering changes only the element coverage, not the message-passing architecture, so the backbone used here is the MACE-MatPES model specialized to the chemistry of the training system.

Variant-specific FiLM settings are summarized in Table~\ref{tab:hparams}. The four FiLM ablations fix the adapter hidden width at $h = 64$ for direct comparison at matched adapter capacity. The headline E-MACE configuration uses $h = 128$ to test whether modest extra capacity in the gating MLP helps. All other settings match Table~\ref{tab:shared_hparams}.

\begin{table}[ht]
\centering
\caption{Variant-specific FiLM settings. All other settings match Table~\ref{tab:shared_hparams}.}
\begin{tabular}{lcc}
\toprule
Variant & FiLM hidden $h$ & FiLM flags \\
\midrule
No-FiLM baseline                   & -    & \texttt{none}                              \\
Concat charge embedding            & -    & \texttt{--embedding\_specs} ($16$-dim categorical) \\
FiLM, $\beta$ only ($\gamma$ off)  & $64$  & \texttt{--charge\_film\_no\_mult}          \\
FiLM, random initialization        & $64$  & \texttt{--charge\_film\_no\_zero\_init}    \\
FiLM, $h=64$                       & $64$  & default                                    \\
\textbf{FiLM, $h=128$ (E-MACE)}    & $128$ & default                                    \\
\bottomrule
\end{tabular}
\label{tab:hparams}
\end{table}

The four per-state MACE-MatPES specialists used in the cross-evaluation matrix (Appendix~\ref{app:crossmatrix}) are fine-tuned from the same MACE-MatPES backbone with the shared settings of Table~\ref{tab:shared_hparams}, but with each specialist seeing only the AIMD frames at its single training charge $q\in\{0,6e,10e,16e\}$. No FiLM block is attached. The NequIP \citep{batzner2022nequip} and Allegro \citep{musaelian2023allegro} baselines in Figure~\ref{fig:model_comparison} are trained from scratch on the same combined four-charge training corpus using each architecture's reference training pipeline.

\section{Cross-pipeline portability of the learned charge response}
\label{app:crossdft}

To test whether the charge-dependent response E-MACE learns from its VASP r$^2$SCAN training labels generalizes across DFT engines, we evaluate the same model on the same atomic configurations using GPAW \citep{enkovaara2010gpaw} in plane-wave mode at a $500$ eV cutoff with PAW pseudopotentials and PBE exchange-correlation. The model has never seen any GPAW or PBE label at any charge during training. The geometry pool is a fixed set of neutral water configurations sampled from the $q=0$ AIMD trajectory. For each charge $q\in\{2e,4e,\ldots,20e\}$ the response is measured on this same pool both ways. GPAW settings: $\Gamma$-point sampling on the $324$-atom cells (edge $13.8$ to $14.8$ \AA{}). Gaussian smearing (Methfessel-Paxton order 0) with width $0.001$ eV at $q=0$ (effectively closed-shell insulator) and $0.05$ eV at non-zero charges.

\begin{figure}[ht]
\centering
\includegraphics[width=0.8\linewidth]{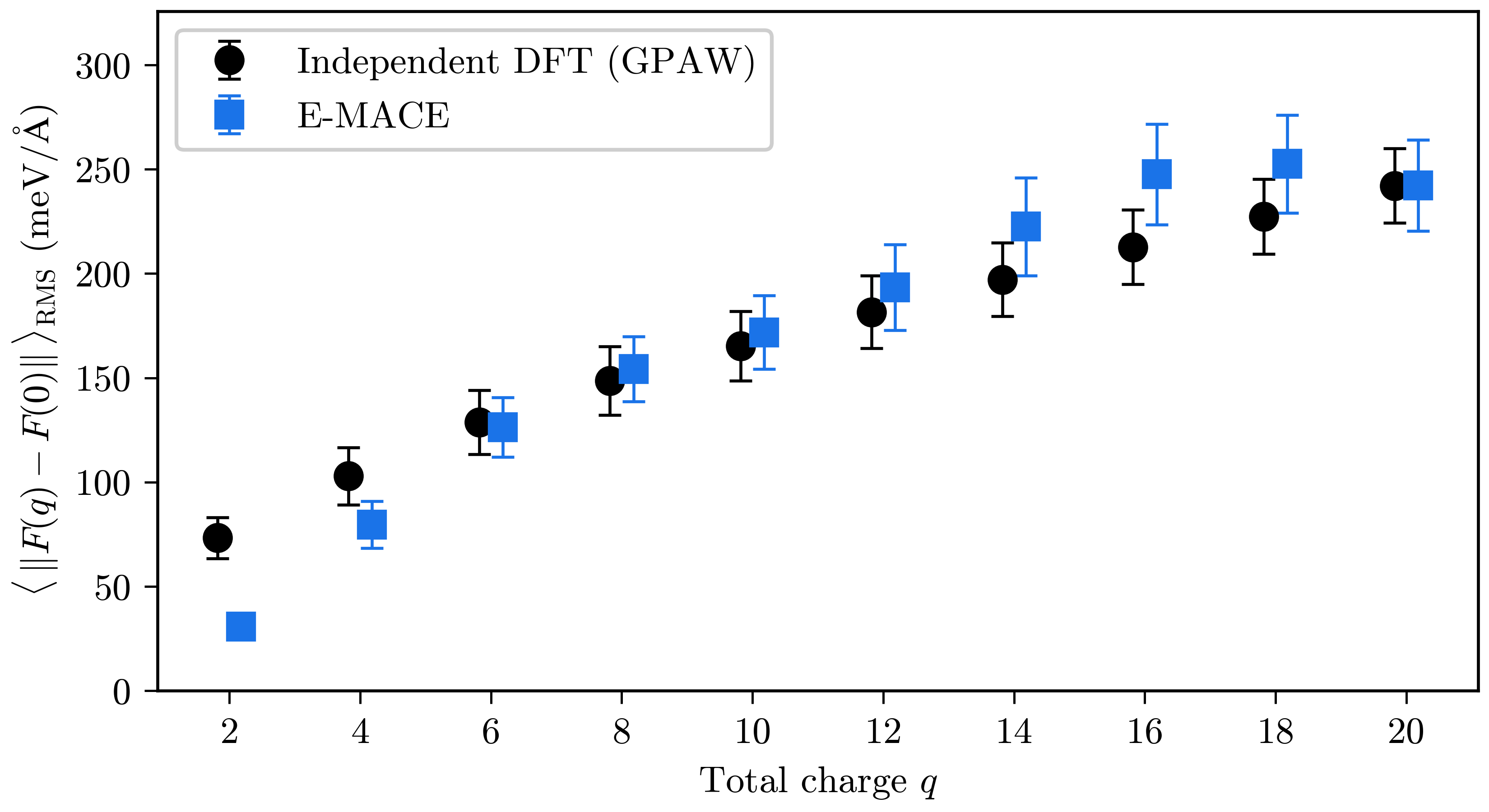}
\caption{Cross-pipeline per-atom force-difference response on a fixed pool of neutral water geometries. Black: independent DFT reference (GPAW PW $500$ eV, PBE). Blue: E-MACE prediction. Error bars are the geometry-pool standard deviation. The two pipelines agree within those error bars across most of the studied range. At the smallest charges $q=2e$ and $q=4e$, E-MACE underestimates the GPAW reference, while at $q\in\{14e,16e,18e\}$, E-MACE sits slightly above the GPAW reference.}
\label{fig:forcedeltavscharge}
\end{figure}

For each geometry $G_i$ in the pool and each charge $q$, the per-atom RMS magnitude of the force change induced by the conditioning input is computed,
\begin{equation*}
\Delta F_{\rm RMS}(q;\,G_i)
\;=\;
\Big\langle \big\| \mathbf F(q;\,G_i) - \mathbf F(0;\,G_i) \big\| \Big\rangle_{\rm RMS\ over\ atoms},
\end{equation*}
both with E-MACE (predicting on $G_i$ first at conditioning value $q$ and then at $q=0$) and with two independent GPAW self-consistent calculations on $G_i$ (one at total charge $q=0$, one at $q$). Figure~\ref{fig:forcedeltavscharge} reports the geometry-pool mean and standard deviation of $\Delta F_{\rm RMS}$ at each charge, overlaid for the two pipelines.

Across the bulk of the studied range, $q\in\{6e,8e,\ldots,20e\}$, the two pipelines agree within geometry-pool error bars, with the response magnitude rising smoothly from $\approx 130$ meV/\AA{}{} at $q=6e$ to $\approx 250$ meV/\AA{} at $q=20e$. At the smallest charges, $q=2e$ and $q=4e$, E-MACE underestimates the GPAW reference: at $q=2e$ E-MACE predicts $\approx 30$ meV/\AA{} against a GPAW value of $\approx 73$ meV/\AA{}, a shortfall of roughly $60\%$, and at $q=4e$ the shortfall is roughly $25\%$. At $q\in\{14e,16e,18e\}$ E-MACE sits slightly above the GPAW reference, by roughly $10$ to $15\%$, with overlapping $1\sigma$ error bars. At $q=20e$ the two pipelines coincide. We hypothesize that both the small-charge shortfall and the intermediate-to-high-charge overshoot are consistent with inter-functional differences between the r$^2$SCAN density functional used to generate the training labels and the GPAW PBE reference. We cannot, however, fully separate this from a genuine model limitation at small charge: the training charges are all $\geq 6e$, so the $q=2e,4e$ regime is the least-supported by the training data, and the excess electrons are most prone to localize there. Such a localized low-$q$ response would be under-predicted by an r$^2$SCAN model trained only on the delocalized high-$q$ regime. We therefore do not claim the discrepancy is purely a functional difference.

This experiment varies the engine (VASP $\to$ GPAW) and the functional (r$^2$SCAN $\to$ PBE) simultaneously, so the agreement we report is across both axes rather than isolating either. That E-MACE matches the GPAW PBE reference on the per-atom force-difference response across more than an order of magnitude in conditioning value, having seen no label from that engine or functional at any charge during training, indicates that the conditioning channel has captured a charge-response that transfers across DFT pipelines rather than fitting engine- or functional-specific artifacts of the training labels.

\section{NVE energy conservation protocol}
\label{app:nveconservation}

In the canonical (NVT) ensemble the particle number, volume and temperature are held fixed, with a thermostat exchanging heat with the system. In the microcanonical (NVE) ensemble the thermostat is removed and the total energy of the isolated system is an exact constant of motion, so any drift in the total energy exposes non-conservative forces. For each studied charge the $324$-atom training cell (cubic, edge $14.8$ \AA{}) is initialized at $300$ K with Maxwell-Boltzmann velocities, then run $0.5$ ps of NVT Langevin dynamics at $300$ K with friction $\gamma_{\rm L}=50$ ps$^{-1}$, using velocity-Verlet integration at a $0.5$ fs time step. The thermostat is then detached and the system continues for $1.5$ ps in pure NVE while $E_{\rm pot}$, $E_{\rm kin}$, and $E_{\rm tot}$ are recorded at every step. All forces are produced by the headline E-MACE configuration with a single set of weights. The interface takes the total charge $q$, which the model converts internally to the conditioning input $c = q/N$, the per-atom excess electron density, with $N=324$ the cell atom count.

Figure~\ref{fig:md_stability} of the main text shows the $1.5$ ps post-detach NVE segment at $q=12e$ (interpolation, top) and $q=20e$ (extrapolation, bottom). The mean total energy drift over the window is $+0.004$ meV/atom at $q=12e$ and $+0.010$ meV/atom at $q=20e$. The bounded $E_{\rm tot}$ band visible in Figure~\ref{fig:md_stability}, approximately $0.18$ meV/atom wide, is the precision floor of the velocity-Verlet integrator at the chosen time step rather than drift: $E_{\rm tot}$ oscillates around a fixed mean rather than walking away from it. A non-conservative force field would instead produce a sloped trace (linear drift accumulating over the window), not a centered band. The centered band is observed uniformly across all studied charges. Because the FiLM block depends only on the per-graph scalar $c$ and acts only on scalar channels, the conditioning input does not introduce non-conservative pathways into the predicted energy, and forces remain exact gradients of a position-dependent energy uniformly across $q$.

\section{Pair correlation analysis}
\label{app:pdf}

For each charge $q\in\{0,4e,8e,14e,18e\}$ the $324$-atom training cell (cubic, edge $14.8$ \AA{}) is propagated under NVT Langevin dynamics at $300$ K with friction $\gamma_{\rm L}=50$ ps$^{-1}$ and a $0.5$ fs time step. Pair correlations are accumulated from a multi-picosecond production segment following a $0.5$ ps equilibration.

Each partial pair correlation $g_{\alpha\beta}(r)$ is histogrammed with a $0.02$ \AA{} bin width and Gaussian-smoothed in $r$ with $\sigma=0.10$ \AA{}. In the single $324$-atom cell, $g(r)$ is rigorously defined only out to the minimum-image limit $r\le L/2 = 7.4$ \AA{}. For the difference signal $\Delta g_{\alpha\beta}(r;q)$ the shell-normalization and finite-size contributions largely cancel between the two charges, but we restrict attention to $r\leq 7$ \AA{} where the partial correlations are well resolved. Experimental PDF resolution ($\sigma\approx 0.2-0.3$ \AA{} at a reciprocal-space cutoff of $\approx 12$ \AA{}$^{-1}$) would broaden these features further while preserving the charge trends. The reduced PDF used in the body,
\begin{equation*}
G(r;q) \;=\; 4\pi r\, \rho_0\,\bigl[g_{\rm tot}(r;q) - 1\bigr],
\end{equation*}
takes $g_{\rm tot}(r;q)$ as a weighted sum of the three partial correlations $g_{\rm OO}, g_{\rm OH}, g_{\rm HH}$ with weights set by the relevant atomic scattering factors. For water under either X-ray or ultrafast electron probes the O-O correlation dominates this sum.

\begin{figure}[ht]
\centering
\includegraphics[width=0.8\linewidth]{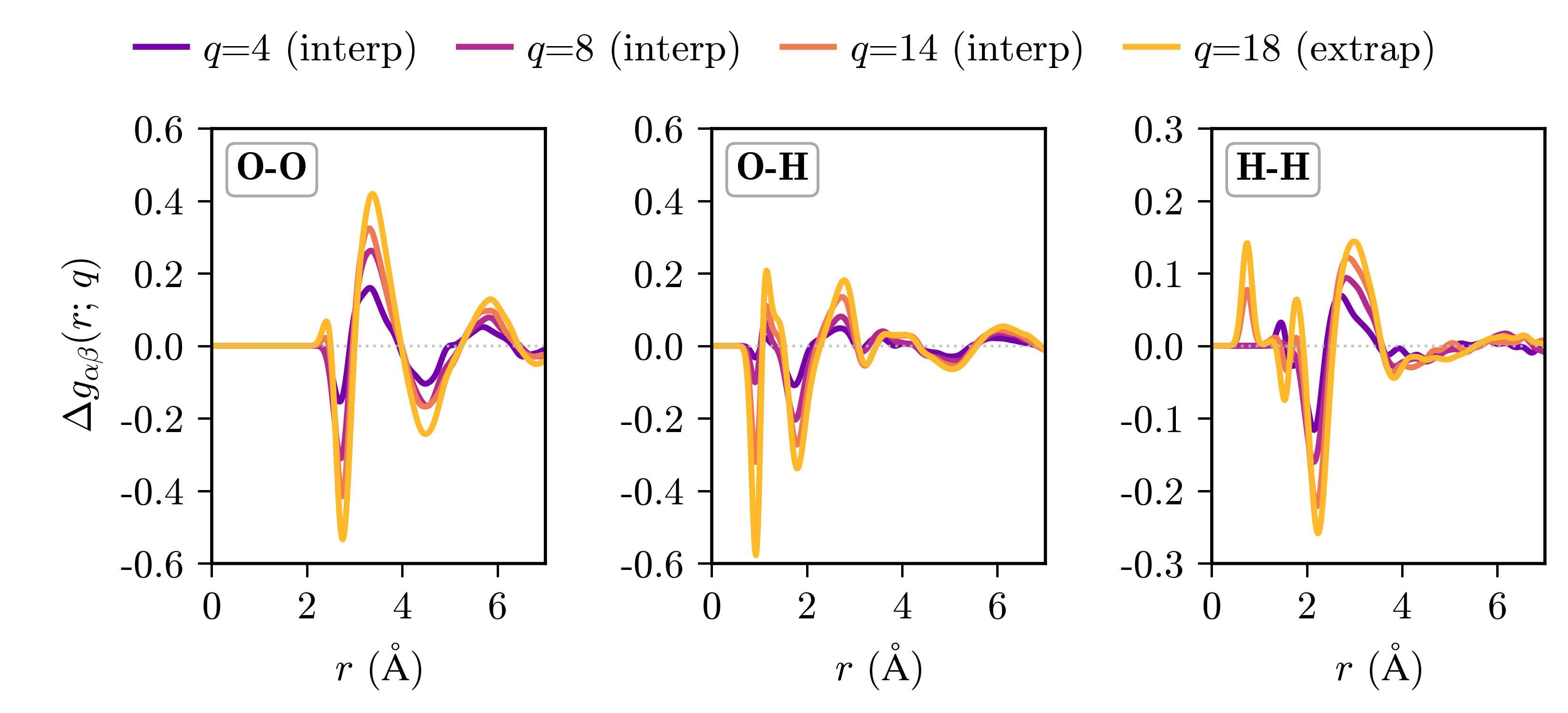}
\caption{Charge-induced structural response $\Delta g_{\alpha\beta}(r;q) = g_{\alpha\beta}(r;q) - g_{\alpha\beta}(r;0)$ for the three pairs O-O (left), O-H (middle), H-H (right) at $q\in\{4e,8e,14e,18e\}$. The first O-O coordination shell shifts outward (loss at $\approx 2.7$ \AA{}, gain at $\approx 3.3$ \AA{}). The O-H and H-H pairs show the same outward redistribution. All features scale monotonically with $|q|$.}
\label{fig:gr_pairs}
\end{figure}

Figure~\ref{fig:gr_pairs} reports the partial structural responses $\Delta g_{\alpha\beta}(r;q) = g_{\alpha\beta}(r;q) - g_{\alpha\beta}(r;0)$ separately for the three pairs at $q\in\{4e,8e,14e,18e\}$. The first O-O coordination shell loses density at $r\approx 2.7$ \AA{} and gains it at $r\approx 3.3$ \AA{} as charge is added, signaling an outward shift of the nearest-neighbor O-O distance. The O-H pair correlation loses weight near the inter-molecular hydrogen bond distance ($r\approx 1.7$ \AA{}) and gains it at $r\approx 2.4$ \AA{}, and the H-H pair shifts outward in the same direction with density redistributing from $r\approx 2.0$ \AA{} to $r\approx 2.8$ \AA{}. The amplitude of every feature grows monotonically from $q=4e$ through the extrapolation charge $q=18e$, with no qualitative change of shape across that range. The reductive H$_2$-formation peak in the H-H partial grows continuously with charge and is present in the $q=16e$ AIMD training frames, confirming a physical process rather than an extrapolation artifact.

\section{Finite-size cross-check in a \texorpdfstring{$2\times 2\times 2$}{2x2x2} supercell}\label{app:supercell}

The structural responses in the main text are computed in the $324$-atom training cell, where the pair correlation is rigorously defined only to the minimum-image limit $r\le 7.4$ \AA{}. To confirm that the charge-induced structure is not a finite-size artifact and survives at larger system size, the analysis is repeated in a $2\times 2\times 2$ supercell ($2,592$ atoms, cubic edge $29.6$ \AA{}, valid to $14.8$ \AA{}). The supercell is seeded by tiling a charge-equilibrated $324$-atom frame rather than the neutral cell. This charge-adapted initial condition is required for MD stability at the higher charges, and even so the highest charges require sampling several Langevin seeds and retaining those that remain stable, because pure microcanonical (NVE) dynamics at the largest extrapolation charge $q=20e$ in the larger cell does not conserve energy. Each surviving NVT trajectory uses a $0.5$ fs time step and Langevin friction in the range $50-150$ ps$^{-1}$, with the total charge passed as $q_{\rm cell} = 8q$, which leaves the conditioning input $c = q_{\rm cell}/2592 = q/324$ identical to training.

\begin{figure}[ht]
\centering
\includegraphics[width=0.8\linewidth]{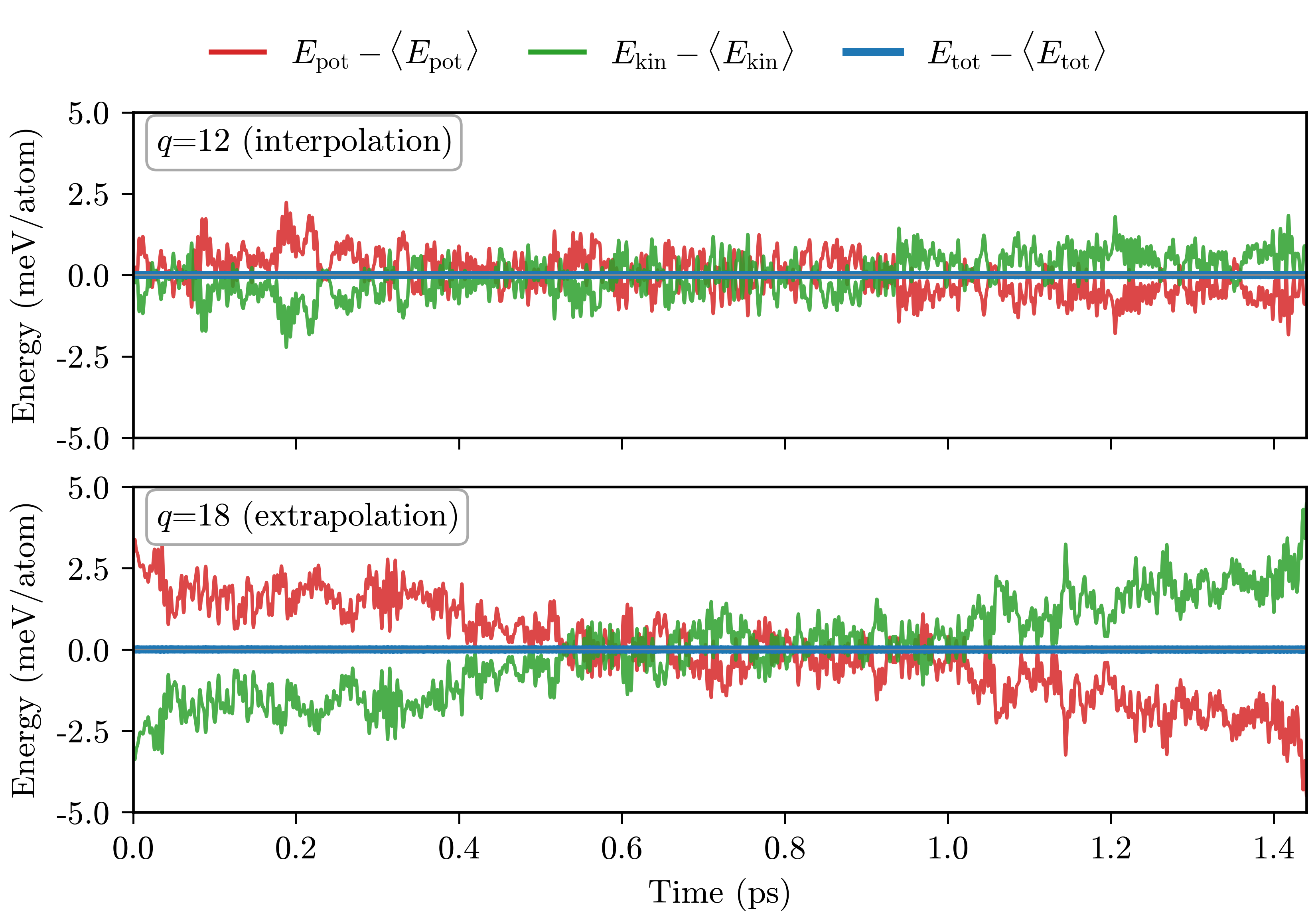}
\caption{Energy conservation in the $2\times 2\times 2$ supercell ($2,592$ atoms) over a $1.4$ ps post-thermostat detach NVE window at $q=12e$ (interpolation, top) and $q=18e$ (extrapolation, bottom). As in the $324$-atom cell (Figure~\ref{fig:md_stability}), $E_{\rm pot}$ (red) and $E_{\rm kin}$ (green) exchange thermal-scale energy while their sum $E_{\rm tot}$ (blue) remains bounded with mean drift below $1$ meV/atom. The per-atom fluctuation amplitude is smaller than in the $324$-atom cell by roughly $1/\sqrt{8}$, as expected from the larger particle number. One set of E-MACE weights produces stable, energy-conserving dynamics across the interpolation-extrapolation range at this system size.}
\label{fig:md_stability_2x2x2}
\end{figure}

The seeds that survive equilibration conserve energy over a multi-picosecond window at $q=12e$ and $q=18e$. Figure~\ref{fig:md_stability_2x2x2} shows post-thermostat detach NVE traces at $q=12e$ (interpolation) and $q=18e$ (extrapolation) in the $2,592$-atom cell: $E_{\rm tot}$ stays bounded with mean drift below $1$ meV/atom over a $1.4$ ps window, reproducing the $324$-atom energy-conservation result (Figure~\ref{fig:md_stability}) at $8\times$ system size.

\begin{figure}[ht]
\centering
\includegraphics[width=0.8\linewidth]{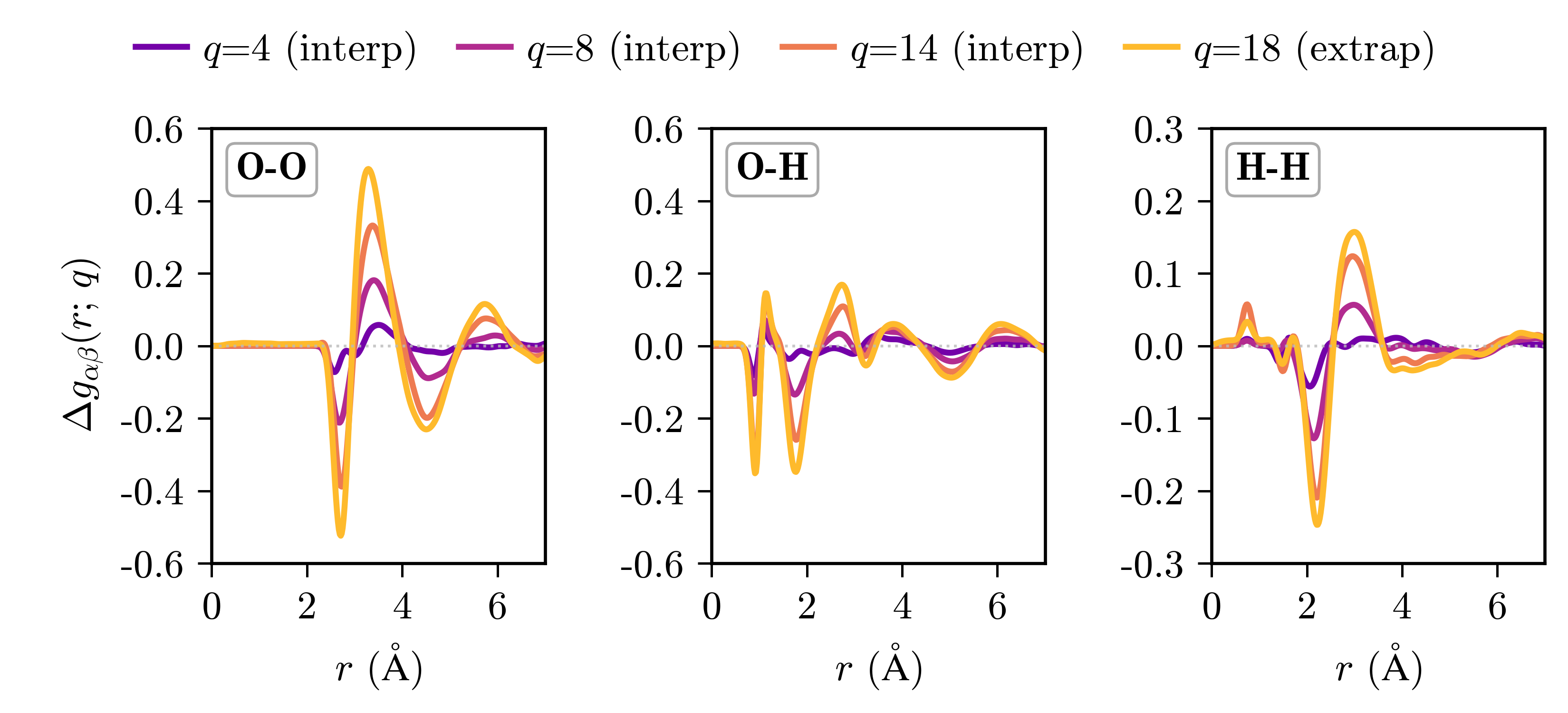}
\caption{Per-pair partial pair-correlation differences $\Delta g_{\alpha\beta}(r;q)$ (O-O, O-H, H-H) in the $2\times 2\times 2$ supercell ($2,592$ atoms, $8\times$ larger than the main-text cell), decomposing the total $\Delta G(r;q)$ of Figure~\ref{fig:supercell_pdf} into its constituent partials. The per-pair trends reproduce the $324$-atom result (Figure~\ref{fig:gr_pairs}) at larger system size.}
\label{fig:supercell_pairs}
\end{figure}

Turning to structure, the per-pair decomposition in the supercell (Figure~\ref{fig:supercell_pairs}) reproduces the $324$-atom partials of Figure~\ref{fig:gr_pairs}: the O-O depletion of the first shell is the dominant signal, with smaller, consistent shifts in the O-H and H-H partials.

\begin{figure}[ht]
\centering
\includegraphics[width=0.8\linewidth]{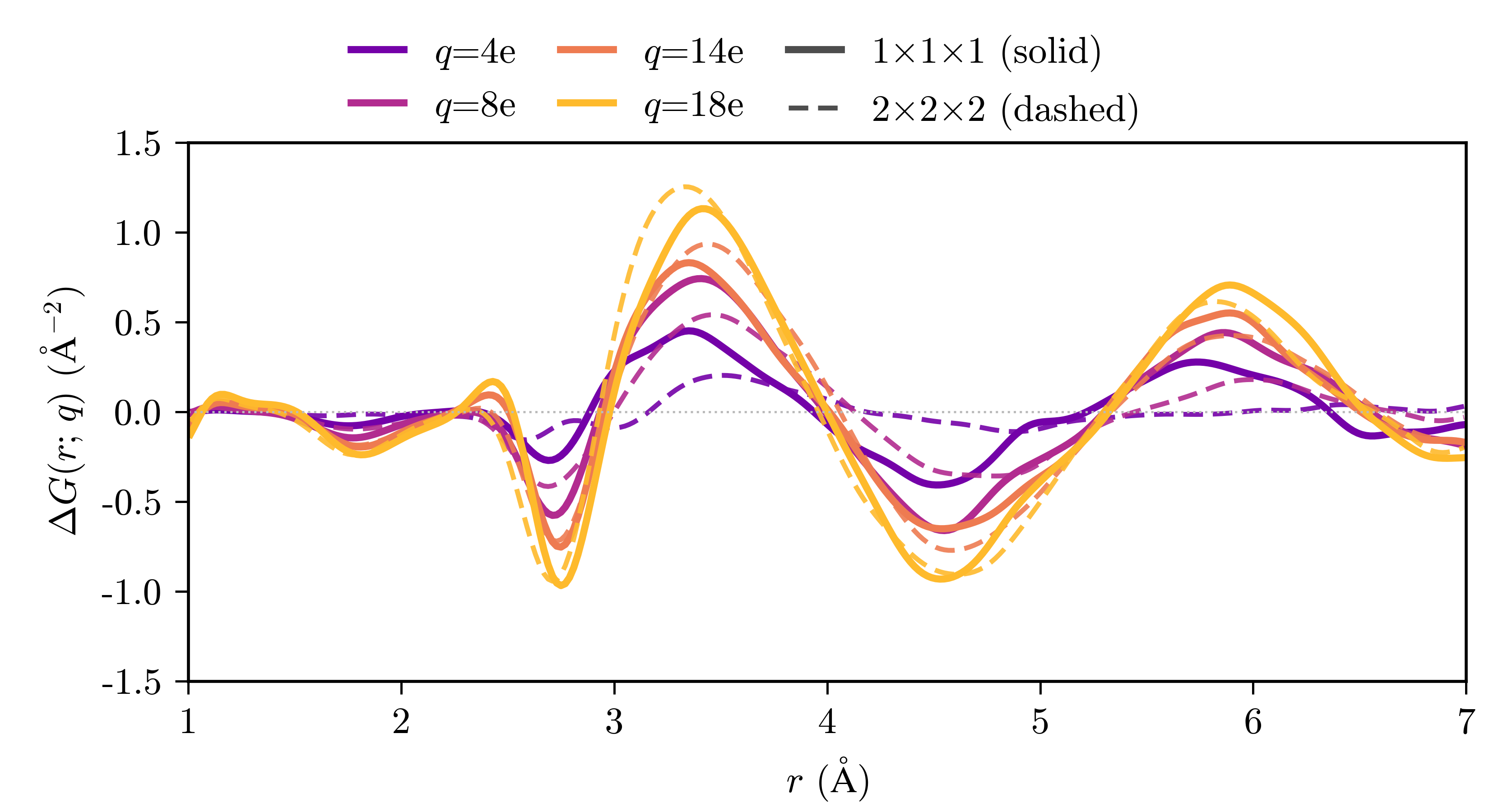}
\caption{Reduced pair-distribution-function difference $\Delta G(r;q)$ at $q\in\{4e,8e,14e,18e\}$, overlaying the $324$-atom $1\times1\times1$ cell (solid) and the $2,592$-atom $2\times2\times2$ supercell (dashed). The two cell sizes track each other closely at every charge, confirming that the charge-induced structural response reproduces at $8\times$ system size and is free of finite-size artifacts.}
\label{fig:supercell_pdf}
\end{figure}

Summing the partials, Figure~\ref{fig:supercell_pdf} shows the reduced PDF difference $\Delta G(r;q)$ in the supercell. The first O-O peak height decreases monotonically with charge in both cells and agrees between them within $\approx 0.07$ across all charges (1$\times$1$\times$1 vs. 2$\times$2$\times$2: $q = 0$, $2.51$ vs. $2.46$; $q = 4e$, $2.37$ vs. $2.44$; $q = 8e$, $2.21$ vs. $2.27$; $q = 14e$, $2.09$ vs. $2.08$; $q = 18e$, $1.97$ vs. $1.97$). The charge-induced structural trend is therefore system-size independent over the range studied, as expected for an equilibrium pair correlation above the correlation length.

\end{document}